\documentclass[10pt,twocolumn,letterpaper]{article}

\usepackage[pagenumbers]{wacv} % To force page numbers, e.g. for an arXiv version

\usepackage{graphicx}
\usepackage{amsmath}
\usepackage{amssymb}
\usepackage{booktabs}

\usepackage{multirow}
\usepackage{xcolor}
\usepackage{bm}
\usepackage[accsupp]{axessibility} % Improves PDF readability for those with disabilities

\usepackage[pagebackref,breaklinks,colorlinks]{hyperref}

\usepackage[capitalize]{cleveref}
\crefname{section}{Sec.}{Secs.}
\Crefname{section}{Section}{Sections}
\Crefname{table}{Table}{Tables}
\crefname{table}{Tab.}{Tabs.}

\def\wacvPaperID{1466}
\def\confName{WACV}
\def\confYear{2025}

\begin{document}

\title{TimberVision: A Multi-Task Dataset and Framework for Log-Component Segmentation and Tracking in Autonomous Forestry Operations}

\author{Daniel Steininger ~~~~ Julia Simon ~~~~ Andreas Trondl ~~~~ Markus Murschitz \\
AIT Austrian Institute of Technology (Center for Vision, Automation \& Control) \\
{\tt\small\{daniel.steininger,julia.simon,andreas.trondl.fl,markus.murschitz\}@ait.ac.at}}

\maketitle

\begin{abstract}
Timber represents an increasingly valuable and versatile resource. However, forestry operations such as harvesting, handling and measuring logs still require substantial human labor in remote environments posing significant safety risks. Progressively automating these tasks has the potential of increasing their efficiency as well as safety, but requires an accurate detection of individual logs as well as live trees and their context. Although initial approaches have been proposed for this challenging application domain, specialized data and algorithms are still too scarce to develop robust solutions. To mitigate this gap, we introduce the TimberVision dataset, consisting of more than 2k annotated RGB images containing a total of 51k trunk components including cut and lateral surfaces, thereby surpassing any existing dataset in this domain in terms of both quantity and detail by a large margin. Based on this data, we conduct a series of ablation experiments for oriented object detection and instance segmentation and evaluate the influence of multiple scene parameters on model performance. We introduce a generic framework to fuse the components detected by our models for both tasks into unified trunk representations. Furthermore, we automatically derive geometric properties and apply multi-object tracking to further enhance robustness. Our detection and tracking approach provides highly descriptive and accurate trunk representations solely from RGB image data, even under challenging environmental conditions. Our solution is suitable for a wide range of application scenarios and can be readily combined with other sensor modalities.
\end{abstract}

\section{Introduction}
\begin{figure}
    \centering
    \includegraphics[width=1.0\columnwidth]{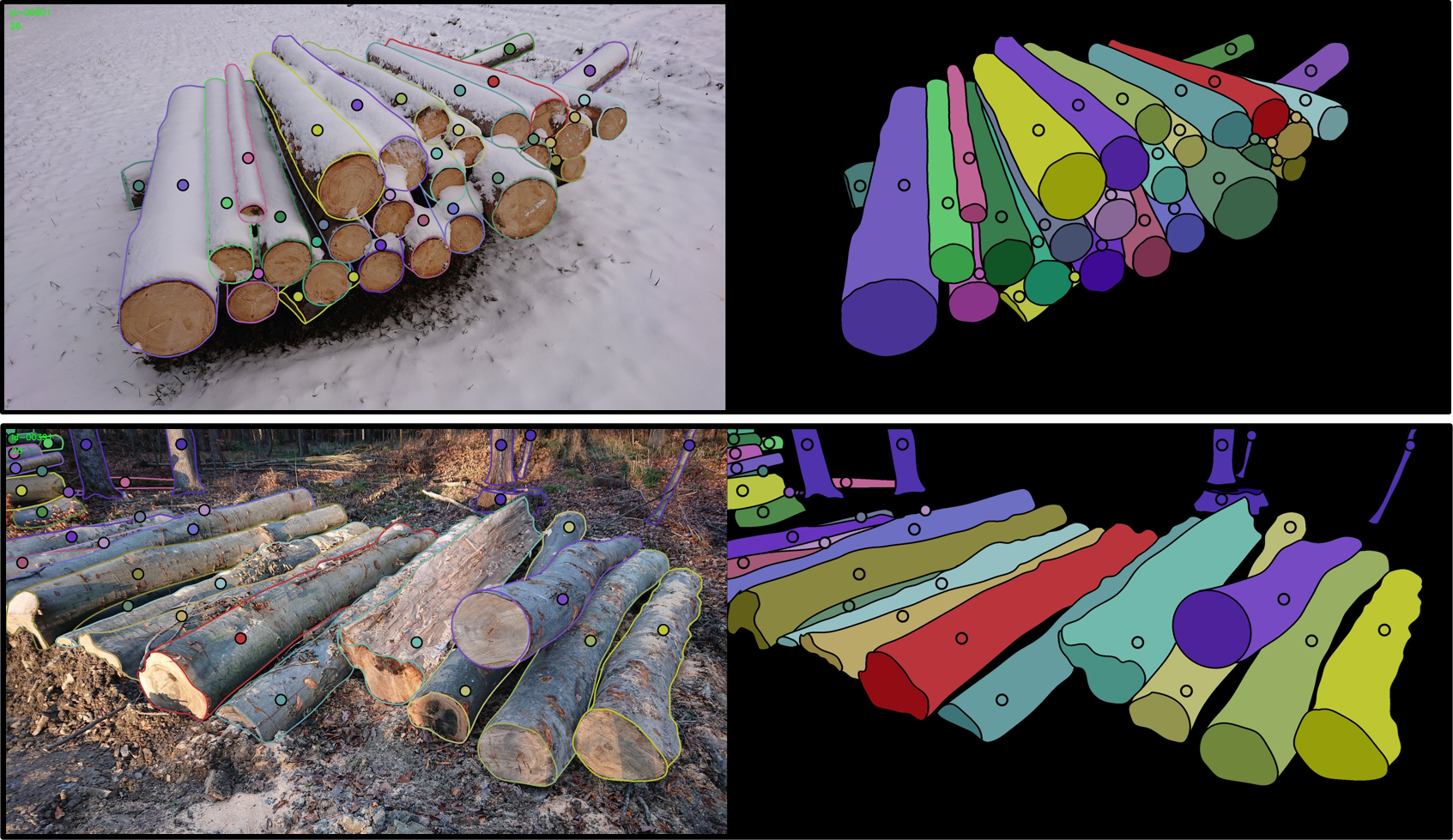}
    \caption{Representative examples of semi-automatically generated annotations for instance segmentation of multiple trunk components in the TimberVision dataset.}
    \label{fig:annotations}
\end{figure}

Wood has been a valuable resource since prehistoric times, gaining further importance since the industrial age due to its multitude of uses. Today, it is used not only for furniture, building materials and paper, but also fabric, insulation, bio-gasoline and many other products. While the processing of wood is often highly automated, the early stages of its value chain still require significant human labor and involve considerable risk, as forestry workers are among the group with the highest fatality rates per full-time employee \cite{Us22}. Automating operations such as cutting, processing and handling logs as well as forest inventory holds the potential to increase both safety and efficiency. Several works have focused on automating harvesters and forwarders, but due to the high complexity of the tasks, these systems either operate only in ideal testing scenarios \cite{LaPe24} or experience dangerous failures due to the lack of visual input data \cite{JeJu21, RoSc09}. In our opinion, these limitations mostly result from the lack of a modern, scalable and efficient machine-learning solution for identifying logs along with the training data it requires. Our objective is to develop a dataset and approach addressing the autonomous or assisted harvesting, loading and measuring of logs, i.e. cut stems and branches. A detection approach has to provide not only their position but ideally also their orientation and contour as well as the positions of their cut and lateral surfaces, as visualized by the annotation examples in \cref{fig:annotations}. This information is essential for deriving geometric cues such as middle axis and boundaries. We aim to infer these features exclusively from RGB data to provide an easily accessible solution, which can be used on its own or combined with 3D information to estimate the center of gravity required for precisely grasping and therefore efficiently manipulating each log.

Learning-based vision approaches are considered a key technology for autonomous operations in unstructured environments. However, covering all application scenarios from natural forests to more ordered saw-mill environments requires significant efforts in acquiring and preparing relevant image data, as well as a specialized approach for training and fusing multiple learning tasks. To address these research objectives, we propose the following contributions:
\begin{itemize}
    \item
        We provide a novel multi-task image dataset and efficient annotation pipeline focusing on logs and their components in various application scenarios along with multiple annotated scene parameters.\footnote{Dataset, code and models are available for academic use at 
        \href{https://github.com/timbervision/timbervision}{https://github.com/timbervision/timbervision}} 
    \item 
        We conduct comprehensive ablation experiments for real-time-capable oriented object detection (OOD) and instance segmentation (ISEG) and analyze the impact of scene configurations on model performance.
    \item 
        We introduce an extensible framework for fusing both learning tasks, correlating individual components into unified object representations, deriving their geometric information and tracking them over time.
\end{itemize}
\section{Related work}
The degree of automation in forestry operations is still low, partially due to the lack of training and test data capturing the required variability for robustly detecting the organic shapes of tree trunks in this highly unstructured environment. Some existing approaches heavily rely on 3D data captured by LIDAR sensors, either using model-based cylinder fitting \cite{MaGu24, PrRa22} or learning-based approaches \cite{WaCh19, ChNa20, KrTa21} for detecting live trees. While they perform well for inventory applications in uncluttered forests, they are usually not applicable for log piles, dead trees or dense vegetation. Similarly, satellite or aerial imagery \cite{FiBe22, FiCa23} does not provide the perspective required for handling individual logs. Synthetically generating data, as in \cite{GrPo22}, holds the advantage of high quantity, but requires a lot of effort to bridge the gap to real imagery. The most readily available data are ground-based RGB images. There are multiple navigation-focused datasets containing trees, either in the context of urban scenarios \cite{JoBr21, XiLi19, LuDe21, QiYa24, ChLi22} or unstructured outdoor environments \cite{WiEu19, JiOs21, LaLe23}. While all of them provide segmentation masks, they, too, provide only live trees as one of multiple classes. The number of datasets exclusively focusing on the detection of tree trunks in RGB images is very limited. The ForTrunkDet dataset \cite{SiQu21} provides axis-aligned bounding-box annotations for live trees on 2.9k RGB and thermal images, but does not contain segmentation masks or cut logs. CanaTree100 \cite{GrFo23} contains 100 images with segmentation masks for 920 live trees. The TimberSeg dataset \cite{FoGa22}, on the other hand, focuses exclusively on cut logs, omitting live trees. It consists of 220 images with segmentation masks for 2500 trunks and is therefore the most closely related to our application domain. However, in images with log piles, the annotation only includes their top layer, thereby limiting the applicability for tasks such as log counting, volume estimation or handling multiple logs at once. None of the existing datasets differentiates between individual trunk components such as cut and lateral surfaces or includes both live and cut trees. Furthermore, there are few datasets, even in other domains, incorporating individual meta-parameters, such as daytime \cite{SaDa19} or weather \cite{MiMi19}, and even fewer providing a comprehensive analysis of data variability and model robustness \cite{ZeHo18, StKr22}. Our TimberVision dataset incorporates all these features and additionally contains the largest number of instance-segmentation masks in real RGB images in the domain of forestry operations. 

Object detection is one of the first learning tasks addressed by many established learning-based architectures \cite{LiMa14, ReDi16, LiGo17, ChWa19}. While most of them use axis-aligned bounding boxes, recent works such as DOTA \cite{XiBa18, XiCh21} show the advantages of oriented bounding boxes (OBBs) for elongated objects, which provide a significantly better representation for detecting and tracking cut logs with arbitrary poses. Similarly, pixel-wise class assignment can be conducted using semantic segmentation \cite{LoSh15, FiKo17}, but the more recent development of instance segmentation \cite{HeGk17} provides separated contours of individual objects and is therefore more relevant for our purpose. Panoptic segmentation \cite{KiHe19, LiZh21} could provide the most descriptive representation, but requires a full annotation of all visible classes, most of which are not relevant for our application. For both selected tasks, there are approaches based on convolutional neural networks \cite{WuKi19, JoCh23} as well as vision transformers \cite{ZhSu20, StGa21}, with the former still providing competitive accuracy and the advantages of faster inference along with more efficient deployment.
\section{The TimberVision dataset}
To address the data gap in the domain of forestry operations, we gathered a comprehensive set of relevant images from multiple sources and designed a process to efficiently derive scene parameters and annotations, as detailed below.

\subsection{Image acquisition}
Since the amount of publicly available image data suitable for tree-trunk detection is limited, we recorded our own data in two stages. Initially, we focused on forests accessible by public transportation and hiking, mainly around Vienna, but also including other areas of Austria, Slovakia and the Czech Republic. We captured about 3k images with a total of eleven sensors (including smartphones, an SLR camera and a UAV) on more than 40 separate days between December 2021 and June 2024, thereby covering a wide range of environmental and lighting conditions as well as seasonal effects. Resolutions range from 1280x720 to 3000x2000 pixels with a median of 2016x1512. Each day typically includes multiple sites containing different numbers and arrangements of logs as well as varying backgrounds. Later on, we complemented the data with images recorded by a ZED 2 camera while operating a crane throughout different loading and harvesting scenarios. While the first setup is designed to maximize the variety of log appearances and environmental conditions, the latter is more closely related to specific application scenarios. In both cases, we verified that the published images do not contain any sensitive personal information.

\subsection{Scene parameters}
\label{sec:scene_parameters}

\begin{figure}
    \centering
    \includegraphics[width=1\columnwidth]{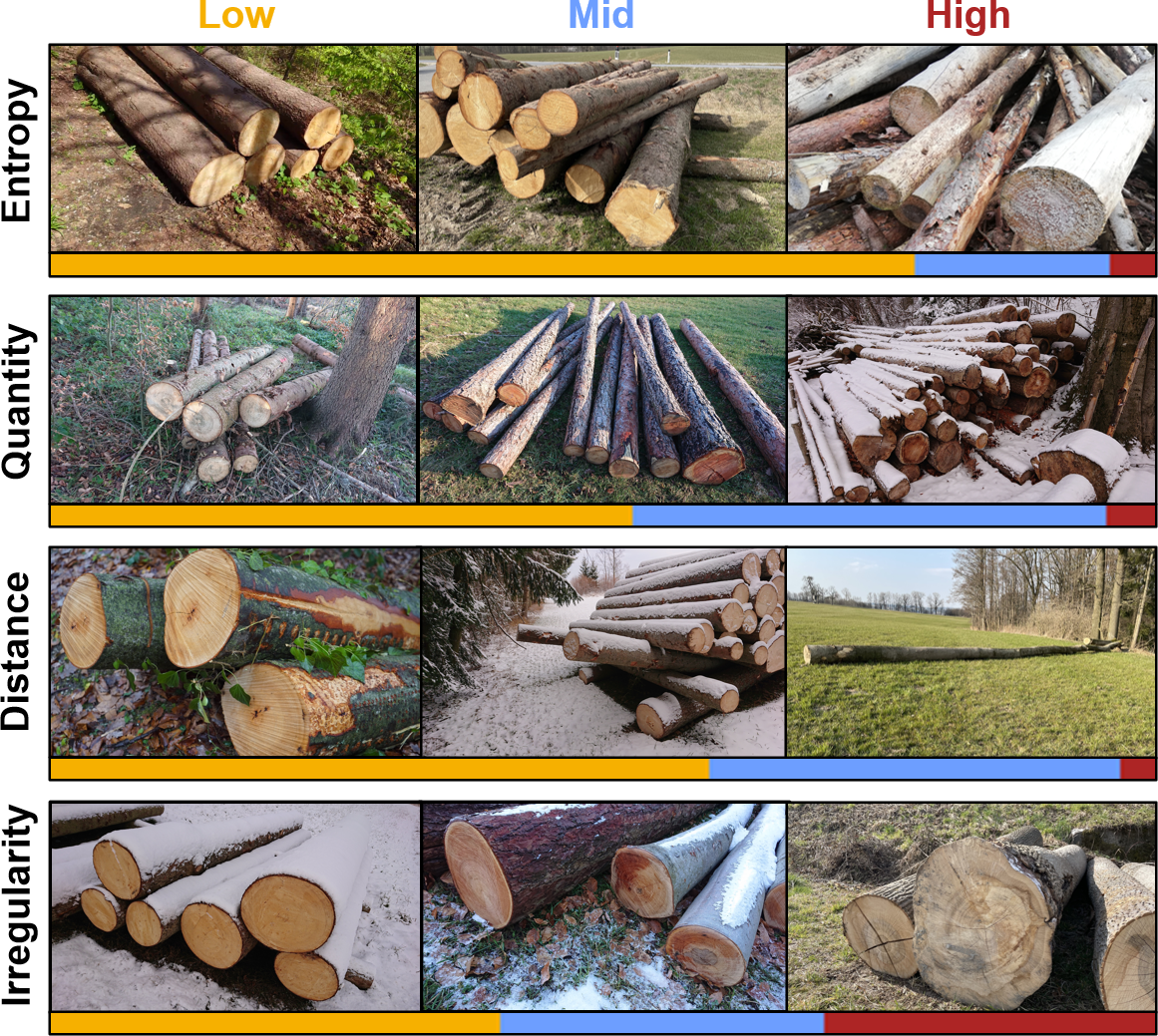}
    \caption{Representative image samples for \textcolor[HTML]{f6b000}{Low}, \textcolor[HTML]{6d9eff}{Mid} and \textcolor[HTML]{ad2525}{High} intensities of annotated scene parameters. The color bars show their distributions across the dataset.}
    \label{fig:scene_parameters}
\end{figure}

Efficiently choosing a representative but feasible set of samples for further annotation and later evaluating the performance of trained models in specific scenarios both require a high-level description of scene properties and context. Beside a binary tag for images containing snow, we defined four parameters and assigned them one of three intensities for each image, as visualized in \cref{fig:scene_parameters}:

\begin{itemize}
    \item
        \textbf{Entropy} describes the arrangement of logs varying from parallel stacks to unstructured wood piles.
    \item
        \textbf{Quantity} defines the number of depicted trunks, with 8 and 30 marking the minimum counts for \textit{Mid} and \textit{High} settings, respectively.
    \item
        \textbf{Distance} roughly categorizes the offset between the majority of trunks and the camera, thereby indicating the number of truncated logs at image borders. Images completely filled with logs are labeled as \textit{Low}, while \textit{High} indicates the visibility of entire instances.
    \item
        \textbf{Irregularity} refers to the shape of trunks, ranging from approximately cylindrical to highly uneven.       
\end{itemize}

\subsection{Instance annotation}
\label{sec:instance_annotation}
\begin{figure}
    \centering
    \includegraphics[width=1\columnwidth]{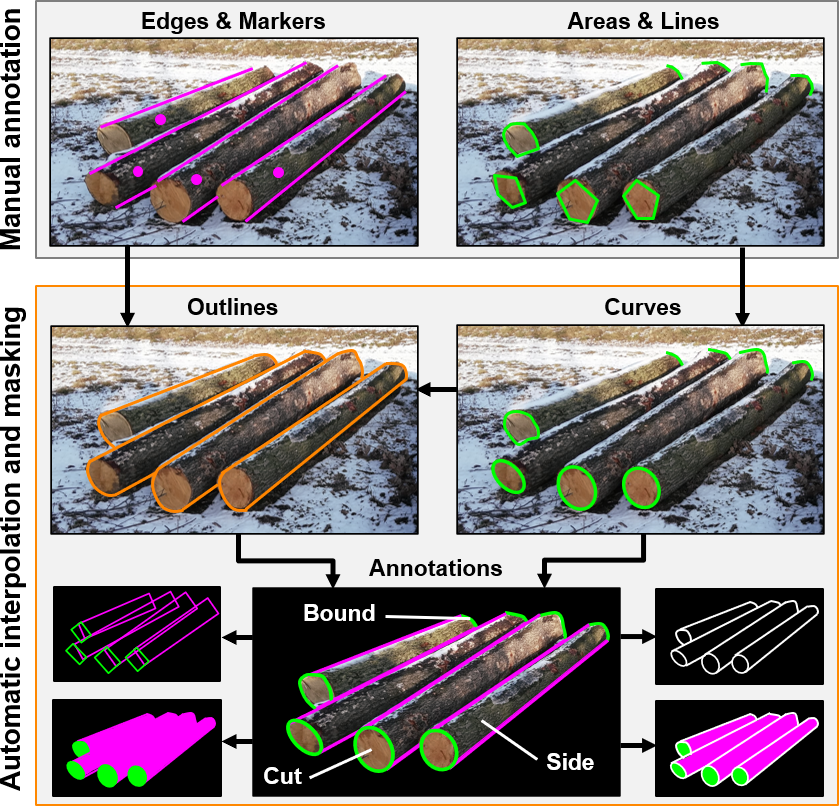}
        \caption{Overview of our annotation pipeline automatically deriving multi-task annotations from manual point annotations.}
    \label{fig:annotation_pipeline}
\end{figure}

Since annotations are conducted in-house, we aim to derive them for a maximum of different learning tasks with as little manual effort as possible. With open-source tools as the preferred solution, Scalabel's \cite{Sc24} lane-detection setup turned out to be a good fit for our requirements. As visualized in \cref{fig:annotation_pipeline}, it allows the definition of trunk components, including their surface discontinuities, as poly-lines based on only a few selected points. Each instance is assigned a label differentiating between lateral \textit{Edges} separating logs from the each other or the background, visible \textit{Section Areas} as well as \textit{Section Lines} denoting the visible borders of cross-sections facing the other way. Additional points unambiguously denote areas covered by each trunk between its constituent lines. Fitting all markers and lines facilitates annotation of instances down to a width of 8 pixels. As a final step, each component is assigned an ID unique to the trunk instance it belongs to or ID 0 for live trees and rooted stumps. For selected sequences, IDs are set consistent over time to allow the evaluation of multi-object tracking.

Based on these manual point annotations, detailed two-dimensional representations for multiple tasks are derived automatically. Annotated \textit{Edges} and \textit{Section Lines} are interpolated as smooth splines, while \textit{Section Areas} are used for fitting either ellipses for regular logs or closed splines for more complex forms. These boundaries combined with the area markers form comprehensive representations of three types of trunk components, which can be converted into arbitrary formats. \textit{Side} surfaces represent only the lateral area of trunks, \textit{Cut} surfaces their cross sections and \textit{Bound}aries their back-facing sections. \textit{Trunks} can include multiple of these components. As a final step, the generated annotations are verified by human annotators. In the future, we plan to use our own results or foundation models such as Segment Anything \cite{KiMi23} or CLIPSeg \cite{LuEc22} for generating pre-annotations and thereby further increasing efficiency.

\subsection{Dataset statistics}
\label{sec:dataset_statistics}
We annotated more than 2k images containing about 26k trunks categorized into six subsets by their origin and depicted scenarios, as listed in \cref{tab:dataset_overview}. \textit{Core} contains the images captured in forests and other outdoor locations. It is complemented by extensions consisting of \textit{Loading} and \textit{Harvesting} scenarios with visible machinery and third-party \textit{OpenSource} data. \textit{Tracking} consists of keyframes evenly sampled from video sequences at 2 FPS and annotated with consistent object IDs over time.

\begin{table}
    \begin{center}
        \begin{tabular}{l|rrrr}
            \textbf{Subset} & \textbf{Rec} & \textbf{Images} & \textbf{Trunks} & \textbf{Comp} \\
            \hline
            \textbf{Core} & 71 & 1,415 & 17,999 & 33,445 \\
            \textbf{Loading} & 18 & 220 & 3,853 & 8,782 \\
            \textbf{Harvesting} & 10 & 45 & 825 & 1,720 \\
            \textbf{OpenSource} & 42 & 42 & 776 & 1,634 \\
            \hline
            \textit{\textbf{Tracking}} & \textit{13} & \textit{266} & \textit{1,900} & \textit{3,708} \\
            \textit{\textbf{TimberSeg*}} & \textit{10} & \textit{35} & \textit{940} & \textit{2,049} \\
            \hline
            & \textbf{164} & \textbf{2,023} & \textbf{26,293} & \textbf{51,338} \\
        \end{tabular}
    \end{center}
    \caption{TimberVision dataset statistics. \textbf{Images} are clustered into \textbf{Rec}ording sessions based on timestamps and GPS tags, ensuring a difference of at least two minutes and one hundred meters between them. \textbf{Trunks} can consist of up to three \textbf{Comp}onent classes.}
    \label{tab:dataset_overview}
\end{table}

\cref{tab:dataset_comparison} relates our dataset to the most similar existing works. The most relevant synthetic dataset is SynthTree43k. However, like CanaTree100, it focuses on live trees, which are only a secondary target of our work. The closest match in terms of scenarios and input modalities is TimberSeg, which, however, provides only a fraction of the image and instance quantities in TimberVision and less detailed annotations, as only the top layers of log piles and no individual trunk components are included. To evaluate the generalization of our models, we created new annotations following our policy for a representative selection of TimberSeg images (denoted as \textit{TimberSeg*} in \cref{tab:dataset_overview}). More details regarding annotation compatibility and dataset statistics can be found in the supplementary material. To the best of our knowledge, TimberVision surpasses all current real-image trunk-segmentation datasets in terms of quantity as well as annotation depth by a large margin.

\begin{table}
    \begin{center}
        \begin{tabular}{l|cccccc}
            \textbf{Dataset} & \textbf{Images} & \textbf{Trunks} & \textbf{C} & \textbf{R} & \textbf{L}\\
            \hline
            SynthTree43k \cite{GrPo22} & 43k & 162k & - & - & - \\
            CanaTree100 \cite{GrFo23} & 100 & 920 & - & \checkmark & - \\
            TimberSeg \cite{FoGa22} & 220 & 2.5k & - & \checkmark & \checkmark \\
            \textbf{TimberVision} & \textbf{2k} & \textbf{26.3k} & \checkmark & \checkmark & \checkmark \\
        \end{tabular}
    \end{center}
    \caption{Comparison of TimberVision to existing forestry datasets. Beside the numbers of annotated \textbf{Images} and \textbf{Trunks}, we rate the inclusion of \textbf{C}omponent annotations, \textbf{R}eal rather than synthetic images and cut \textbf{L}ogs.}
    \label{tab:dataset_comparison}
\end{table}
\section{Methodology}
Based on our dataset, we perform comprehensive ablation experiments for the tasks of OOD and ISEG. To combine their advantages, we demonstrate a generic approach for fusing their output into unified trunk representations and subsequently tracking them.

\begin{figure*}
    \centering
    \includegraphics[width=2.07\columnwidth]{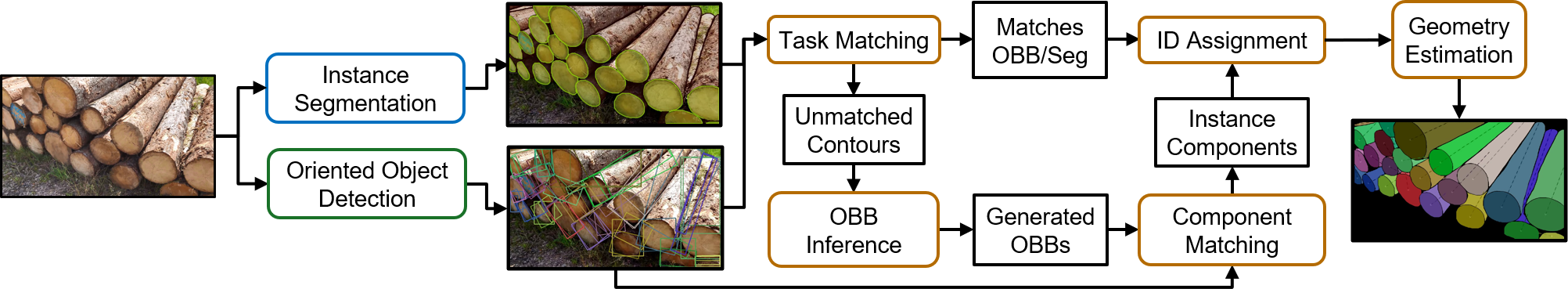}
    \caption{Overview of our task-fusion algorithm deriving unified trunk representations from OOD and ISEG outputs for individual component classes. Processing steps are denoted by rounded and intermediate results by square boxes.}
    \label{fig:task_fusion}
\end{figure*}

\paragraph{Learning experiments}
We build our experiments on the YOLOv8.2 framework \cite{JoCh23}, which provides an established pipeline for training OOD and ISEG, as well as real-time multi-object trackers. We use a random dataset split of 70/15/15 (train/val/test) applied separately for each subset in \cref{tab:dataset_overview}. To avoid overlapping areas between splits, we ensure that images of the same recording session are in one split. The \textit{TimberSeg*} and \textit{Tracking} images are excluded from training and exclusively used for testing generalization and tracking performance, respectively.\footnote{The test set as well as a part of the \textit{Tracking} set are not publicly available to ensure fair benchmarking of community results.} For our cross-dataset fine-tuning experiments based on TimberSeg and CanaTree100, we use five-fold validation to compensate for lower image quantities with random 70/15/15 splits and the original provided cross-validation folds, respectively. Experiments are conducted on an NVIDIA Titan RTX using YOLOv8.2 pre-trained models for initialization and adaptive batch sizes. The remaining hyper-parameters as well as data augmentation adhere to the default settings, except for a reduced weighting of distribution focal loss of 0.01, as this was found to improve convergence. The best models are selected after training for 500 epochs. 

\paragraph{Task fusion and multi-object tracking}
\label{sec:task_fusion}
Our generic approach for fusing OOD and ISEG results of multiple components into robust unified object representations is presented in \cref{fig:task_fusion} and demonstrated for the classes of our dataset. It consists of two matching steps for identifying correspondences between both model outputs in the form of OBBs and segmentation contours, as well as between different components belonging to the same object. Both are conducted using linear sum assignment and differ only by the distance metric to be minimized. For task matching, we generate OBBs enveloping each contour and use their overlap with detected ones. Component matching is based solely on OBBs (either detected or generated) to maximize efficiency and requires dedicated metrics for each class combination. For associating \textit{Cuts} and \textit{Sides}, for instance, we use the largest relative overlap between any of the \textit{Side}'s OBB lines with the \textit{Cut}'s area. Combining all correspondences and eliminating unmatched detections results in unified objects representations consisting of up to one instance of each component represented by an OBB and an optional contour each. For the resulting \textit{Trunks} without both \textit{Bound} and \textit{Cut} components, geometric information is inferred from the remaining classes' relative poses and dimensions, ensuring they are assigned exactly two endpoints.

To preserve object identities over time and increase robustness, we integrate the ByteTrack \cite{ZhSu22} and Bot-SORT \cite{AhOr22} implementations of YOLOv8.2 into our framework. Instead of raw detections, we track new OBBs for unified \textit{Trunks} enveloping all their components, thereby preserving geometric properties and optimizing real-time performance.

\section{Experimental results}
This chapter describes the evaluation of our ablation and learning experiments, including a comprehensive analysis of relevant impact factors on model performance, as well as fusion and tracking results. 

\paragraph{Evaluation protocol and metrics}
\label{sec:metrics}
Regarding our learning experiments and task-fusion algorithm, results are reported solely on the test set using the same input resolutions as during training and an empirically derived confidence threshold of 0.4. We report the challenging \textit{mAP\textsuperscript{50-95}} metric, introduced by the MS COCO benchmark \cite{LiMa14}, which averages multiple Intersection-over-Union (\textit{IoU}) thresholds, as opposed to the still widely used \textit{mAP\textsuperscript{50}} metric \cite{EvVa10} yielding higher, but less descriptive results based on a single threshold. If not stated otherwise, \textit{mAP} scores for ISEG are reported for masks only and not for their less relevant axis-aligned bounding boxes. For fusion results, the same metric becomes even more challenging as it is applied on overall OBBs of unified trunk representations, which only fit the ground-truth if all individual components are accurately detected as well as correctly matched to each other. Since fusion \textit{mAP} scores are therefore not directly comparable to individual model results with multiple classes, we provide additional context in the form of precision and recall.

\begin{figure*}
    \centering
    \includegraphics[width=2.07\columnwidth]{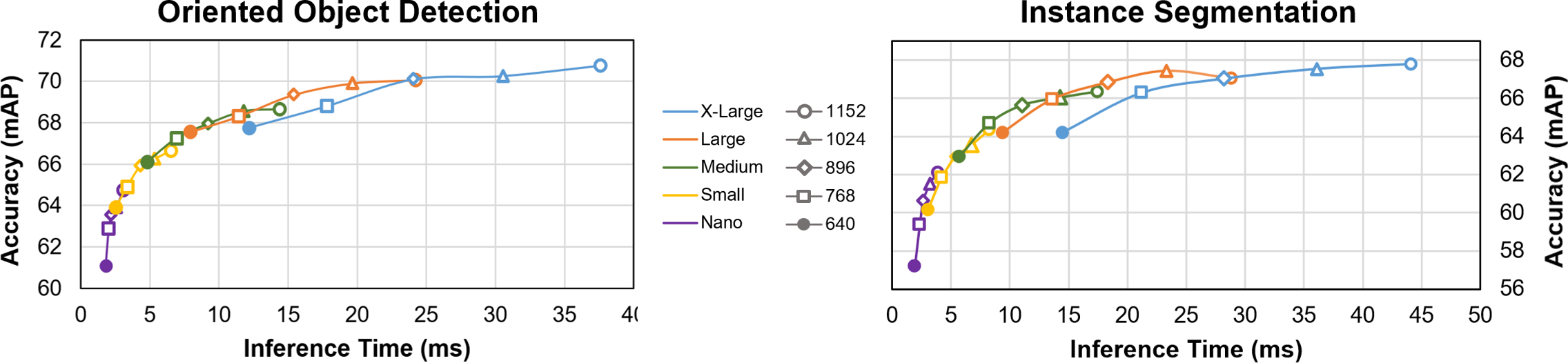}
    \caption{Accuracy as mean class \textit{mAP\textsuperscript{50-95}} and average inference time on test set for multiple model capacities and image sizes.}
    \label{fig:model_ablation}
\end{figure*} 

Tracking performance is evaluated on the dedicated \textit{Tracking} subset using the established py-motmetrics \cite{Py24}. In this case, we explicitly take into account that each trunk can consist of multiple components and rate both segmentation performance and their assignment to trunks by implementing a custom component-wise Intersection-over-Union (\textit{IoU\textsubscript{c}}) score for object masks.
\begin{equation}
IoU_c(G,P)=\frac{\sum_{l\epsilon L}|G_l\cap P_l|}{\sum_{l\epsilon L}|G_l\cup P_l|},
\end{equation}
where \(G_l\) and \(P_l\) are the ground-truth and prediction masks for component \(l\), and \(L\) is the set of available components (i.e. a subset of the labels \textit{Side}, \textit{Cut} and \textit{Bound} in this case). To quantify all aspects of tracking based on this matching score, we discuss Multi-Object-Tracking Accuracy (\textit{MOTA}) \cite{BeSt08} as well as \textit{IDF1} Score \cite{RiSo16}. Furthermore, we report ID Precision (\textit{IDP}), ID Recall (\textit{IDR}) and mean \textit{IoU\textsubscript{c}} over all correct detections (\textit{mIoU\textsubscript{c}}).

\subsection{Ablation experiments}
\begin{figure}
    \centering
    \includegraphics[width=1.0\columnwidth]{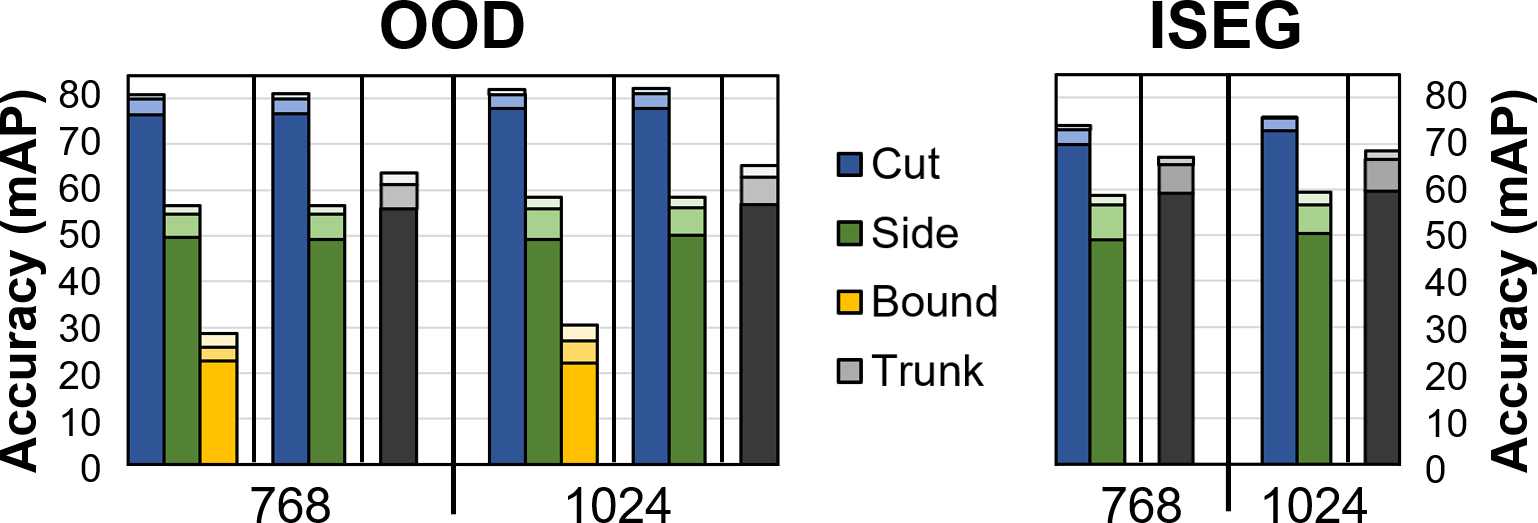}
    \caption{Ablation results as \textit{mAP\textsuperscript{50-95}} on test set for three class combinations and two input sizes. For ISEG, \textit{Bound} is excluded due to its inherent overlap with \textit{Side}. The dark-to-light bar colors denote model capacities \textit{Nano}, \textit{Medium} and \textit{X-Large}, respectively.}
    \label{fig:class_ablation}
\end{figure}

\paragraph{Class ablation} As a preliminary step, we conduct multiple experiments with class combinations ranging from the most fine grained with \textit{Side}, \textit{Cut} and \textit{Bound} as separate classes, to the coarsest combining all of them in a single \textit{Trunk} class.  Results for multiple model variants with different input sizes are presented in \cref{fig:class_ablation} and show consistent trends for both tasks. While \textit{mAP} scores for the \textit{Side} class are slightly lower than those of opaque \textit{Trunk} instances, the values for \textit{Cut} are significantly higher. Since multiple classes also provide more detailed information for post-processing, training two classes is clearly superior to a single-class setup. Another obvious result, however, is the inferior performance of the \textit{Bound} class, which usually covers very small image areas. Following these indications, we rely on a combination of \textit{Cut} and \textit{Side} for subsequent experiments, and infer boundary information geometrically.

\paragraph{Model ablation}
To find the optimal setup, our ablation study includes five image sizes and all standard model capacities, as visible in \cref{fig:model_ablation}. As expected, accuracy and inference time increase with both parameters. For smaller models, accuracy increases stronger than inference time with higher resolutions, while large models show inverse behavior. Based on the analysis, \textit{Large} models with an input size of 1024 pixels provide the best trade-off for our purposes and are used during further evaluation.

\begin{figure*}
    \centering
    \includegraphics[width=1.95\columnwidth]{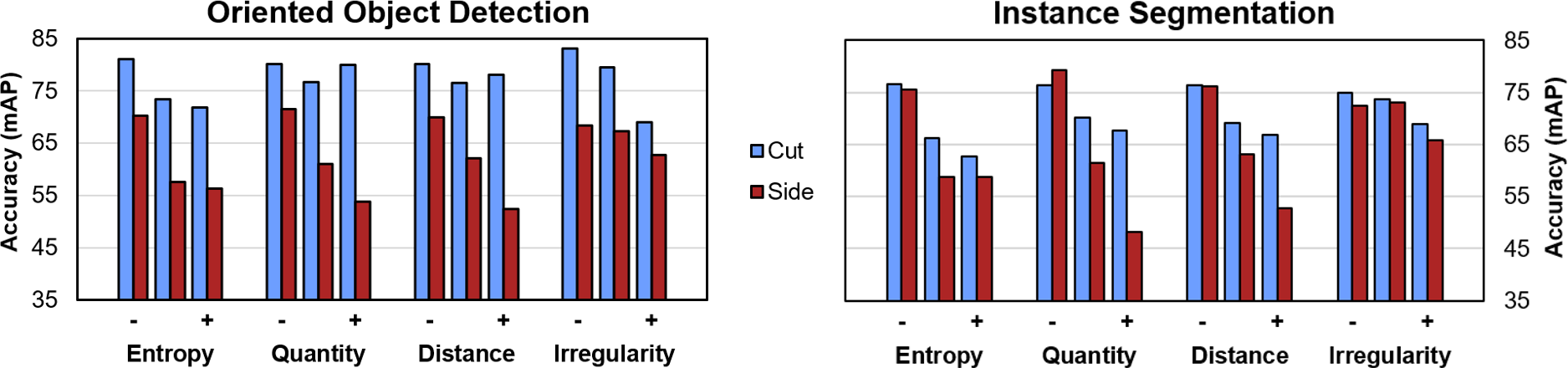}
    \caption{Model performance for different scene-parameter intensities as \textit{mAP} on validation and test splits.}
    \label{fig:performance_impact}
\end{figure*}

\begin{table}
    \begin{center}
        \begin{tabular}{ll|ccc}
            \multicolumn{1}{l}{} &  & \textbf{Base} & \textbf{Tracking} & \textbf{TimberSeg*} \\
            \hline
            \multirow{2}{*}{\textbf{OOD}} & \textbf{Cut} & 81.8 & 80.6 & 67.4 \\
             & \textbf{Side} & 58.0 & 66.0 & 51.7 \\
             \hline
            \multirow{2}{*}{\textbf{ISEG}} & \textbf{Cut} & 76.0 & 73.3 & 56.6 \\
             & \textbf{Side} & 58.9 & 67.3 & 48.4
        \end{tabular}
    \end{center}
    \caption{Model performance as \textit{mAP\textsuperscript{50-95}} for the classes \textit{Cut} and \textit{Side}. Results are reported for \textit{Large} models trained and evaluated on the same image resolution of 1024 pixels on our test (\textit{Base}) and \textit{Tracking} sets and selected images from TimberSeg.}
    \label{tab:model_results}
\end{table}

\subsection{Performance evaluation}
\paragraph{Model performance}
\cref{tab:model_results} reports accuracy on multiple test sets described in \cref{sec:dataset_statistics}. Given the variability of input data, OOD achieves promising results on both of our own test sets, especially for the \textit{Cut} class, which has the most distinct appearance and shape in typical surroundings. Segmentation masks are a more fine-grained representation with slightly lower overall accuracy, but a less distinct gap between the performance of individual classes. Regarding generalization to the images extracted from TimberSeg, our models achieve similar results to the main test set for some scenarios but perform significantly worse for others resulting in overall lower accuracy. Especially nighttime harvesting scenarios with perspectives and tree types highly differing from our recordings, present a limitation for ISEG to be addressed by the next image-acquisition iteration.

To assess the range of practical operating conditions, we evaluate the impact of scene parameters defined in \cref{sec:scene_parameters} on model performance, as summarized in \cref{fig:performance_impact}. As opposed to previous evaluations conducted exclusively on the test set, we now include the validation set as well to ensure a statistically significant number of samples for each parameter. In general, the more visually distinct \textit{Cut} class remains more robust than \textit{Side} against all scene configurations. Especially in case of high trunk quantities which are typically arranged as log piles, \textit{Cuts} tend to be less occluded and well delineated. Regarding tasks, ISEG is better suited for detecting irregularly shaped logs, while OOD is preferable for large quantities of instances. Furthermore, the difference between classes is less pronounced for ISEG, which may be attributed to the larger \textit{Side} instances producing a higher recall than small \textit{Cuts} for this more challenging task. Overall, the results indicate that the models complement each other and perform reasonably well even in challenging scenarios. 

\paragraph{Fusion and tracking}
Overall trunk OBBs generated by our fusion algorithm with the selected models yield an \textit{mAP} of 57.5 on the test set. Given the challenging task of performing both detection and component matching successfully, this score is naturally lower than those for individual models. However, in combination with a high precision and recall of 84.3 and 72.9, respectively, it still indicates that components are rarely matched incorrectly, since a wrong assignment between \textit{Side} and \textit{Cut} instances to the same trunk would drastically distort its overall bounding box. The \textit{mAP} therefore merely suggests that fused trunks might not fit the ground truth as accurately as the results of individual models. This limitation, however, is balanced by the significantly more detailed description of each instance's components and geometric properties.

\begin{table}
    \begin{center}
        \begin{tabular}{l|ccc|cc}
             & \textbf{IDF1} & \textbf{IDP} & \textbf{IDR} & \textbf{MOTA} & \textbf{mIoU\textsubscript{c}} \\
            \hline
            \textbf{ByteTrack} & 71.1 & 85.0 & 61.2 & 63.8 & \textbf{85.8} \\
            \textbf{Bot-SORT} & 72.5 & \textbf{86.6} & 62.4 & 63.7 & 85.4 \\
            \textbf{Optimized} & \textbf{72.9} & 86.0 & \textbf{63.2} & \textbf{65.2} & 85.1 \\
            \hline
            \textbf{Opt} $|$ 10 fps & 66.5 & 81.1 & 56.4 & 57.8 & 84.4 \\
        \end{tabular}
    \end{center}
    \caption{Mean MOT results for default and optimized configurations on all \textit{Tracking} sequences at a default frame rate of 30 fps.}
    \label{tab:mot_eval}
\end{table}

The evaluation of multi-object tracking in \cref{tab:mot_eval} gives a comprehensive impression of detection, component-matching and association quality for our unified \textit{Trunk} instances. Default configurations already yield encouraging results for both \textit{MOTA} and \textit{IDF1} with Bot-SORT showing only marginally superior performance. Precision is higher than recall, suggesting that both delineation and matching of individual trunk components are consistently accurate, but a relevant number of ground-truth samples remains undetected. This is especially true for recordings containing high numbers of thin live trees in the background. Piles of logs partially occluding each other, on the other hand, can be delineated and associated with relatively high precision. Based on these insights, we optimize the Bot-SORT configuration with a lower \textit{new\_track\_thresh} of 0.05 and higher \textit{match\_thresh} of 0.9, which increases \textit{IDF1} and \textit{MOTA} by 0.4 and 1.5, respectively, while also increasing the number of mostly tracked instances according to the Clear-MOT definition from 50.9\% to 52.0\%. Even reducing the detection and tracking frame rate to one third does not drastically decrease performance, further illustrating the application range of our entire pipeline.

\paragraph{Cross-dataset experiments}
\begin{table}
    \begin{center}
        \begin{tabular}{l|cccc|c}
            \multicolumn{1}{c|}{} & \textbf{10} & \textbf{50} & \textbf{100} & \textbf{300} & \textbf{Best} \\
            \hline
            \textbf{TS\textsubscript{COCO}} & 2.0 & 21.6 & 30.5 & 41.3 & \textbf{46.8}\textsubscript{$\pm3.9$} \\
            \textbf{TS\textsubscript{TimberVision}} & 33.2 & 44.7 & 46.5 & 49.4 & \textbf{52.1}\textsubscript{$\pm4.5$} \\
            \hline
            \textbf{CT\textsubscript{COCO}} & 2.5 & 37.8 & 46.8 & 52.9 & \textbf{59.9}\textsubscript{$\pm2.8$} \\
            \textbf{CT\textsubscript{TimberVision}} & 31.7 & 51.3 & 52.8 & 56.5 & \textbf{60.4}\textsubscript{$\pm2.5$}
        \end{tabular}
    \end{center}
    \caption{Evaluation of fine-tuning ISEG models pre-trained with MS COCO and our TimberVision dataset on \textbf{T}imber\textbf{S}eg \cite{FoGa22} and \textbf{C}ana\textbf{T}ree100 \cite{GrFo23} using \textit{Large} models with an input size of 1024. \textit{mAP\textsuperscript{50-95}} is reported on the validation set after different numbers of epochs and on the test set for the \textbf{Best} resulting models along with standard deviations from five-fold validation for the latter.}
    \label{tab:cross_dataset}
\end{table}

To demonstrate the benefits of our dataset and models for other similar domains, we use our final ISEG model as a basis for fine-tuning on both TimberSeg and CanaTree100, which focus on detecting the top layers of log piles and live trees in forest scenes, respectively. Results in comparison to the same approach based on a generic YOLOv8.2 model pre-trained on MS COCO \cite{LiMa14} are summarized in \cref{tab:cross_dataset}. While even overall results on the test set are slightly superior for our model, its main advantage is significantly faster convergence on the new training data. This indicates our models' potential as a basis for fine-tuning on other smaller forestry datasets.

\subsection{Discussion and limitations}
\begin{figure}
    \centering
    \includegraphics[width=1\columnwidth]{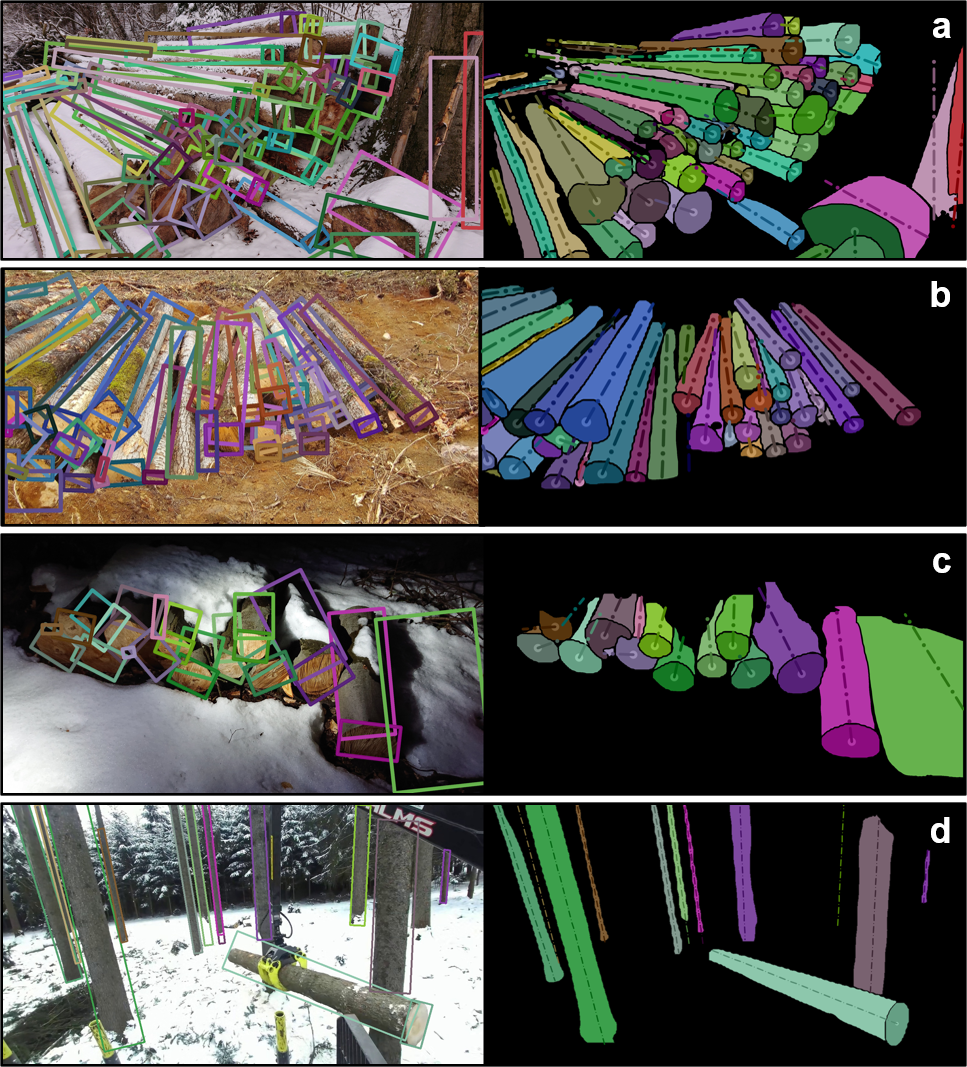}
    \caption{Representative fusion results on test images. The left side shows detected and derived OBBs. The right side shows ISEG results along with center points of cut surfaces and middle axes estimated by our fusion algorithm. Colors identify individual trunk instances, with lighter and darker shades of the same hue corresponding to associated \textit{Side} and \textit{Cut} components, respectively.}
    \label{fig:results}
\end{figure} 

\cref{fig:results} shows a selection of representative fusion results. The combination of OOD and ISEG achieves highly robust detections, even in challenging scenarios, as indicated by the scene-parameter-based evaluation. The approach even generalizes to other datasets such as TimberSeg, as visible in \cref{fig:results}(b). Supporting the quantitative analysis, \textit{Cut} and \textit{Side} components are delineated with high accuracy even if they are covered by snow or debarked, with the visually more distinct \textit{Cut} class usually producing slightly more precise contours. Occlusions are handled reasonably well and only split detections into multiple parts if they affect large areas. Components are correctly assigned to trunk instances in most cases, even for cluttered scenes and large log piles. Furthermore, the results support our thesis that deriving trunk boundaries from inherent geometric properties yields superior performance compared to explicitly detecting them in the form of a \textit{Bound} class, as the middle axes of logs can successfully be inferred for most instances of sufficient visible length. However, there are still a few corner cases to be resolved regarding small or irregularly formed trunks, as visible in the lower right corner of \cref{fig:results}(a). As in this case, wrong component assignment often leads to incorrect estimation of geometric properties, which is the only relevant drawback of this method. 

Overall, false detections are rare and mainly occur in challenging scenarios covered by few training samples such as the combination of darkness and snow in \cref{fig:results}(c). Furthermore, the detection of live trees performs well in the foreground of images, while very thin instances in the far background, as in \cref{fig:results}(d), present a more challenging target and can negatively impact quantitative scores due to their inconsistent detection. However, most practical applications, such as harvesting, loading and inventory tasks, require reliable detections mainly in the near and mid-range, which are comprehensively covered by our approach. More detailed quantitative and qualitative results on all subsets can be found in the supplementary material.

While our results based on the well-established YOLOv8.2 framework introduce a solid baseline for OOD and ISEG, it will be interesting to see the performance of other architectures in the future to identify any potential inherent limitations or biases of this approach. Furthermore, while the incorporation of 3D data would be beneficial for certain applications, we intentionally limited our input data to RGB images to evaluate the potential of this affordable and easily accessible modality by itself before combining it with other sensors.
\section{Conclusion}
Detecting and tracking trunk instances and their components is an essential prerequisite for efficiently handling them during harvesting, loading and measuring tasks. We developed a novel dataset for forestry operations containing annotations for more than 51k log components in real-world images, thereby significantly surpassing all existing similar datasets in terms of quantity and detail. We trained multiple models for the tasks of oriented object detection and instance segmentation and thoroughly evaluated them in different scenarios. Furthermore, we provide a novel fusion and multi-object-tracking framework combining both tasks for real-time perception of trunks and their components as well as geometric properties such as boundaries and middle axes. Our approach detects and delineates trunk components with high accuracy even under challenging conditions, fuses them into unified representations and precisely tracks them across image sequences.

To further improve performance, we plan to integrate additional tracking algorithms and re-identification, as well as to infer segmentation masks directly from oriented instead of axis-aligned bounding boxes. Furthermore, our log representation already includes all relevant information for 3D projection and can therefore be readily combined with depth data to further increase robustness and descriptiveness. By publicly providing the dataset and fusion framework, we aim to extend and develop both in cooperation with the scientific community.

\paragraph{Acknowledgement.}
We would like to thank our colleagues Christian Zinner, Mario Niedermeyer, Verena Widhalm, Marlene Glawischnig and Vanessa Klugsberger for their valuable contributions during image acquisition.

{\small
\bibliographystyle{ieee_fullname}
\bibliography{references}
}

\newpage
\setcounter{page}{1}
\twocolumn[
    \centering
    \Large
    \textbf{TimberVision: A Multi-Task Dataset and Framework for Log-Component Segmentation and Tracking in Autonomous Forestry Operations \\
    \vspace{0.7em}(Supplementary Material)} \\
    \vspace{8.2em}]

\appendix
This supplementary document complements the main paper with additional dataset statistics and presents more detailed results of detection, segmentation and tracking evaluations. Moreover, we illustrate the generalization capacity of our approach to various application domains and conclude with selected corner cases to facilitate a more comprehensive understanding.

\section{Extended dataset description and statistics}
The following sections include detailed statistics and illustrations regarding the process of dataset creation and its final composition.

\subsection{Data acquisition}
\begin{figure}
    \centering
    \includegraphics[width=1\columnwidth]{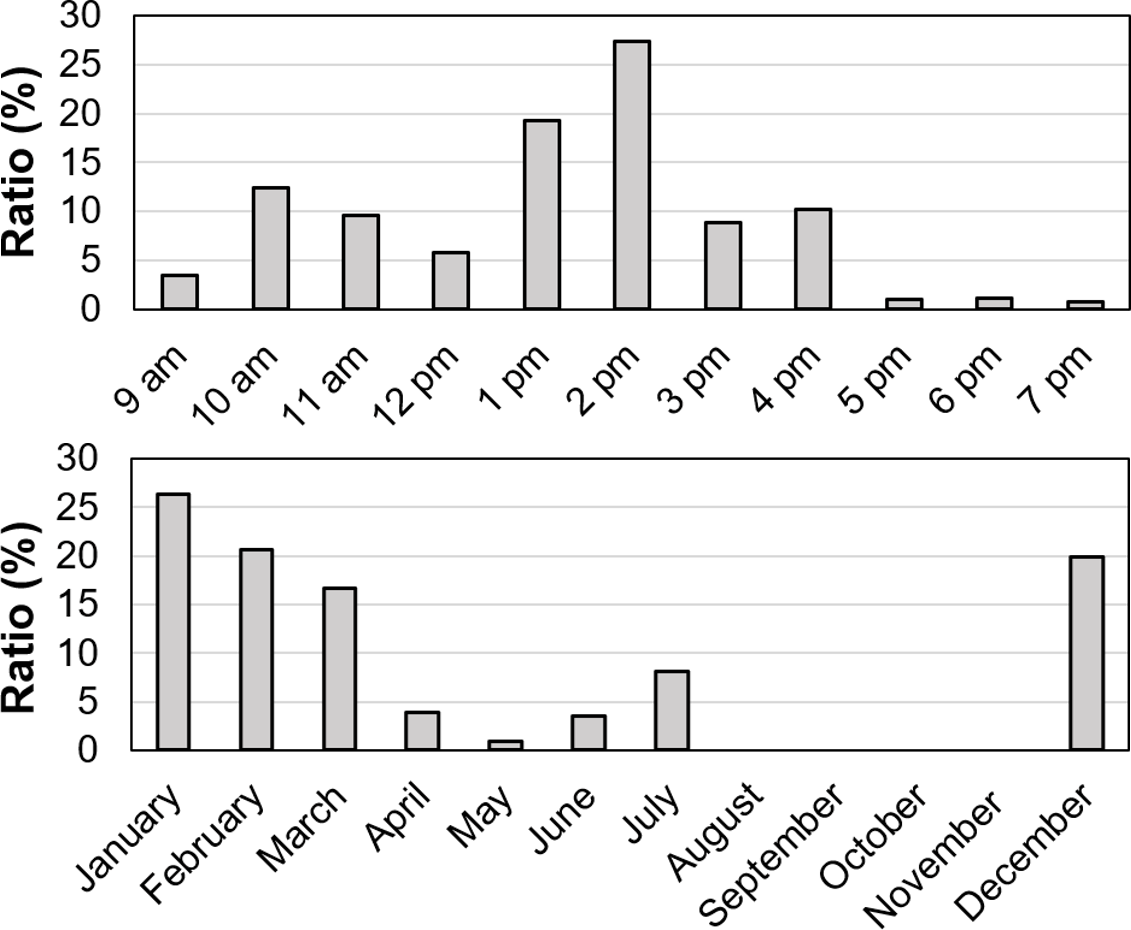}
    \caption{Hourly and monthly recording-time distributions of annotated TimberVision images.}
    \label{fig:distribution_recordings}
\end{figure}

\cref{fig:distribution_recordings} shows the ratios of images captured during specific times of day and months. In 141 recording sessions, we captured a wide variety of seasonal aspects as well as lighting and weather conditions across eight different months. Most data was recorded in winter and early spring, as this is a popular time for harvesting timber. As a result, about 9\% of total images contain snow. Autumn is currently underrepresented and will be the focus of future data campaigns, although winter conditions without snow show partly similar characteristics. In addition to seasonal changes, TimberVision covers a range of daytime variations. Frequent recording times range from morning to late afternoon, while a smaller percentage was captured in the evening. In total, 16 images show dusk or night scenarios.

We captured images for TimberVision using a total of 10 different sensors, which are listed in \cref{tab:sensors} along with their corresponding resolutions and the number of images included in the respective subsets. The additional \textit{OpenSource} subset contains 42 images with resolutions ranging from 672x504 to 3176x2039 pixels. We furthermore recorded image sequences with high scene entropy and log quantities using a DJI Mavic 2 Pro UAV, which were used exclusively for qualitative analysis.

\begin{table*}
    \begin{center}
        \begin{tabular}{lrrrl}
            \textbf{Sensor} & \textbf{Width} & \textbf{Height} & \textbf{Images} & \textbf{Subset} \\
            \hline
            ZED 2 & 1280 & 720 & 304 & Loading, Harvesting, Tracking \\
            Sony Xperia PRO-I & 1280 & 720 & 217 & Tracking \\
            Sony Xperia XZ2 Compact & 1500 & 844 & 8 & Core \\
            Sony Xperia PRO-I & 1920 & 1080 & 30 & Core \\
            Sony Xperia PRO-I & 2016 & 1134 & 29 & Core \\
            Sony Alpha 7S & 2120 & 1192 & 192 & Core \\
            Huawei P20 Lite & 2304 & 1296 & 70 & Core \\
            iPhone 12 Mini & 2016 & 1512 & 279 & Core \\
            Samsung Galaxy S10+ & 2016 & 1512 & 4 & Core \\
            Huawei P20 Lite & 2048 & 1536 & 23 & Core \\
            Blackfly BFS-PGE-31S4C-C & 2048 & 1536 & 10 & Loading \\
            Samsung Galaxy S5 Neo & 2304 & 1728 & 201 & Core \\
            Sony Xperia XZ2 Compact & 2666 & 1500 & 549 & Core, Tracking \\
            Sony Alpha 6000 & 3000 & 2000 & 30 & Core
        \end{tabular}
    \caption{List of sensors used for data acquisition along with their resolutions, numbers of annotated images and associated subsets.}
    \label{tab:sensors}
    \end{center}
\end{table*}

\subsection{Annotation concept}
\begin{figure*}
    \centering
    \includegraphics[width=1.5\columnwidth]{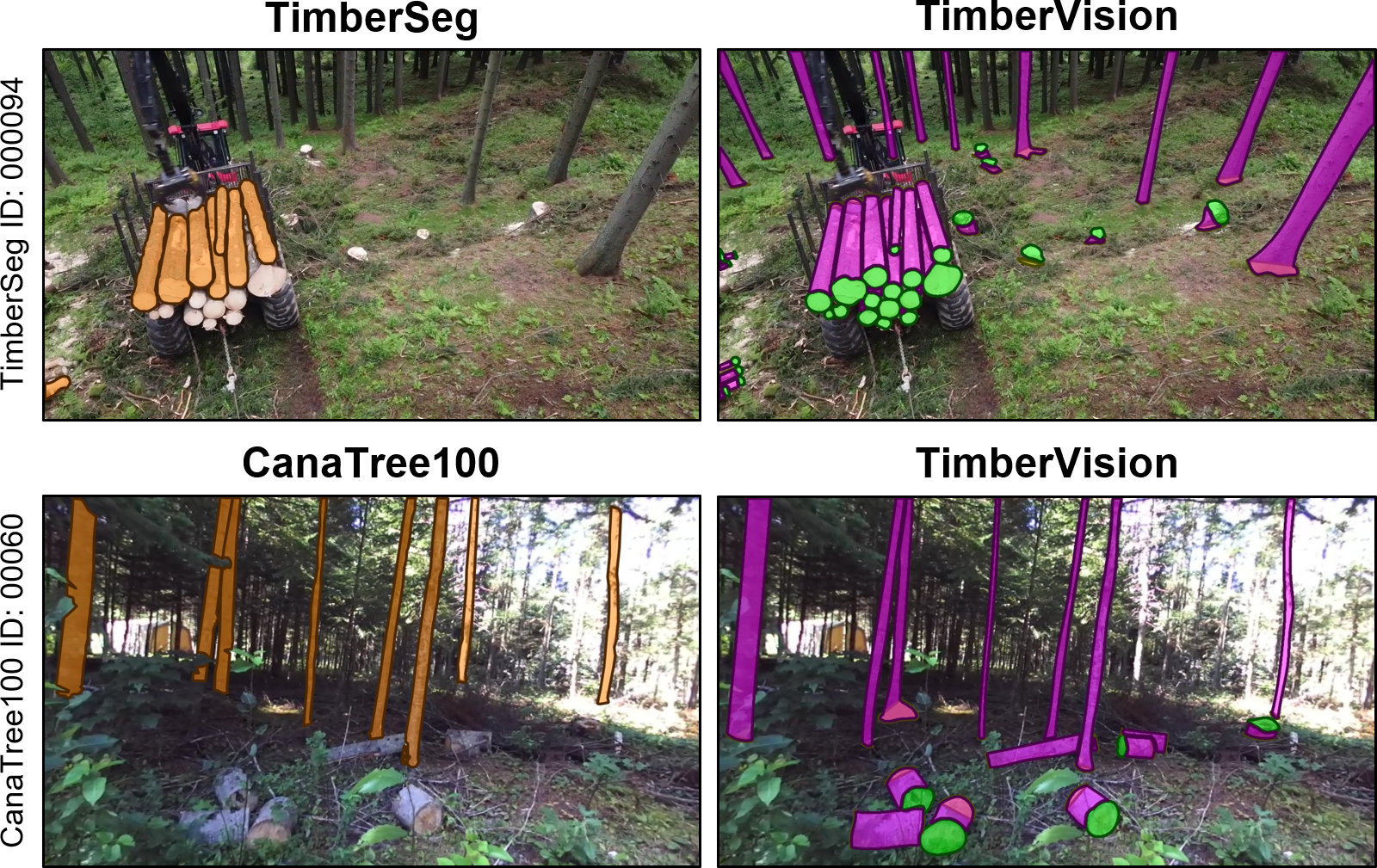}
    \caption{Comparison of annotation schemes in TimberSeg \cite{FoGa22}, CanaTree100 \cite{GrFo23} and TimberVision using original images and corresponding annotations. Orange denotes trunks in single-class annotations, while green and pink identify our \textit{Cut} and \textit{Side} classes, respectively. \textit{Bound} instances overlapping with them are visualized in slightly lighter shades.}
    \label{fig:anno_comparison}
\end{figure*}

\begin{figure}
    \centering
    \includegraphics[width=1\columnwidth]{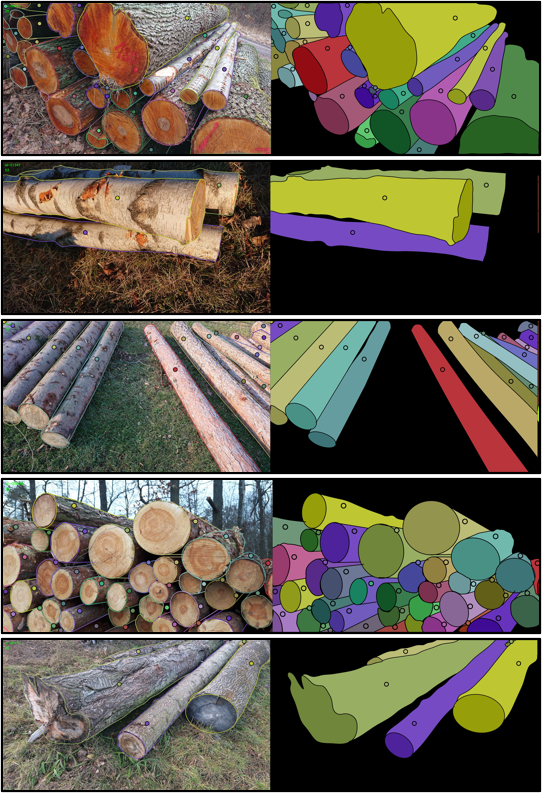}
    \caption{Additional representative examples of semi-automatically generated annotations for instance segmentation of multiple trunk components in the TimberVision dataset.}
    \label{fig:additional_annotations}
\end{figure}

To complement the analysis of other state-of-the-art datasets presented in Tab. 2 of the main paper, \cref{fig:anno_comparison} compares our annotation concept to the approaches of the two most closely related works. Both provide instance-segmentation masks on the level of trunks, while our dataset includes multiple classes for their individual components. TimberSeg focuses on the detection of cut logs and CanaTree100 on live trees, while we include annotations for both types of trunks, separable by their id range. Furthermore, since the target scenario of TimberSeg is mainly log manipulation, only the top layer of log piles is annotated, while we include all visible trunks in a pile, as visible in the top row. Regarding live trees, not all instances in the far background can reasonably be annotated in dense forest scenarios as visible in the second row of images. Our dataset and CanaTree100 set different thresholds for this purpose, which further reduces compatibility, as the same tree may be included in the annotations of one dataset but not the other. Furthermore, the three datasets were recorded at different geographical locations and therefore depict different tree species. Overall, these aspects illustrate the difficulty of comparing the few existing instance-segmentation datasets in the domain of forestry operations or conducting cross-evaluations without substantial limitations or adaptations. On the other hand, the limited compatibility with any existing work shows that our dataset indeed addresses a relevant data gap and represents a valuable complementary addition to the state of the art. To further illustrate our annotation approach, an extended set of representative ground-truth samples is presented in \cref{fig:additional_annotations}.

\subsection{Distribution of scene parameters}
For a more detailed analysis of our training setup, the distribution of scene-parameter intensities across our per-session split between training, validation and test data is shown in \cref{tab:dataset_split}. It demonstrates that combining the validation and test samples in the evaluation of scene-parameter impact in Fig. 7 of the main paper results in a sufficient representation of each intensity. 

\begin{table}
    \begin{center}
        \begin{tabular}{lc|rrr}
            \multicolumn{2}{l|}{\textbf{Scene Parameter}} & \textbf{Train} & \textbf{Val} & \textbf{Test} \\
            \hline
            \multirow{3}{*}{\textbf{Entropy}} & \textbf{-} & 990 & 202 & 175 \\
            & \textbf{} & 190 & 42 & 64 \\
            & \textbf{+} & 23 & 15 & 21 \\
            \hline
            \multirow{3}{*}{\textbf{Quantity}} & \textbf{-} & 634 & 157 & 135 \\
            & \textbf{} & 522 & 91 & 118 \\
            & \textbf{+} & 47 & 11 & 7 \\
            \hline             
            \multirow{3}{*}{\textbf{Distance}} & \textbf{-} & 730 & 164 & 159 \\
            & \textbf{} & 450 & 90 & 92 \\
            & \textbf{+} & 23 & 5 & 9 \\
            \hline
            \multirow{3}{*}{\textbf{Irregularity}} & \textbf{-} & 459 & 107 & 133 \\
            & \textbf{} & 348 & 86 & 63 \\
            & \textbf{+} & 396 & 66 & 64 \\
        \end{tabular}
    \caption{Numbers of images in the training, validation and test splits depicting \textit{Low}, \textit{Mid} and \textit{High} intensities of each annotated scene parameter.}
    \label{tab:dataset_split}
    \end{center}
\end{table}

\begin{figure}
    \centering
    \includegraphics[width=0.85\columnwidth]{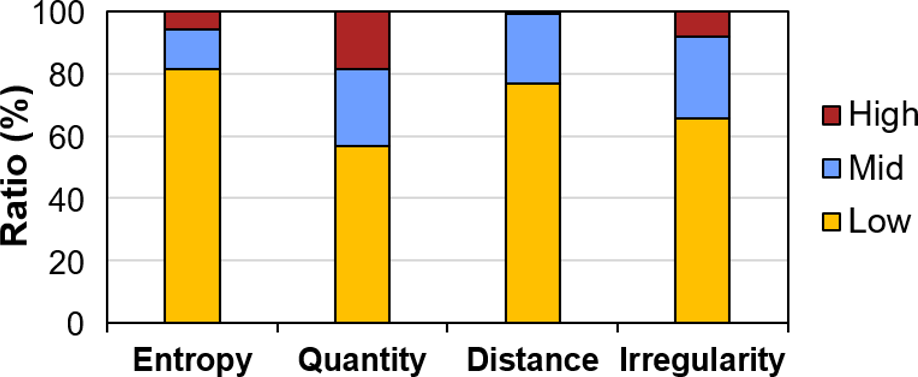}
    \caption{Distribution of scene parameters for annotated keyframes in the \textit{Tracking} subset.}
    \label{fig:meta_tracking}
\end{figure}

To provide more insights about our tracking evaluation and the sequences it is based on, we also summarize the distribution of scene parameters across their keyframes in \cref{fig:meta_tracking}. It shows similar characteristics to the distributions in the overall dataset depicted in Fig. 2 of the main paper and therefore comparable difficulty to the test set used for evaluating detection and fusion performance. 

\subsection{Instance statistics}

\begin{table}
    \begin{center}
        \begin{tabular}{l|rrr}
            \textbf{Subset} & \textbf{Cuts} & \textbf{Sides} & \textbf{Bounds} \\
            \hline
            \textbf{Core} & 9,535 & 17,617 & 6,293 \\
            \textbf{Loading} & 2,522 & 3,589 & 2,671 \\
            \textbf{Harvesting} & 345 & 825 & 550 \\
            \textbf{OpenSource} & 607 & 743 & 284 \\
            \hline
            \textbf{\textit{Tracking}} & 528 & 1,907 & 1,273 \\
            \textbf{\textit{TimberSeg*}} & 425 & 922 & 702 \\
            \hline
            & \textbf{13,962} & \textbf{25,603} & \textbf{11,773} \\
        \end{tabular}
    \caption{Detailed statistics of annotated trunk components in all subsets of the TimberVision dataset.}
    \label{tab:log_components}
    \end{center}
\end{table}

\begin{figure*}
    \centering
    \includegraphics[width=2.07\columnwidth]{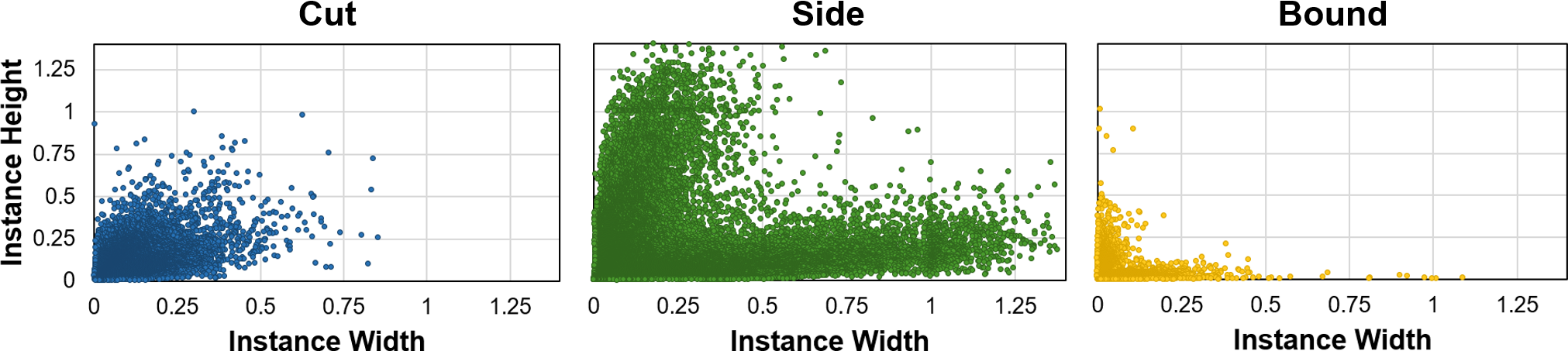}
    \caption{Distribution of instance sizes for each class based on oriented-bounding-box dimensions. Instance width refers to the box side connecting the leftmost corner with the adjacent one in counter-clockwise direction, meaning that a value larger than the corresponding height indicates an orientation below ninety degrees in \cref{fig:orientation}.}
    \label{fig:instance_sizes}
\end{figure*}

\begin{figure}
    \centering
    \includegraphics[width=1\columnwidth]{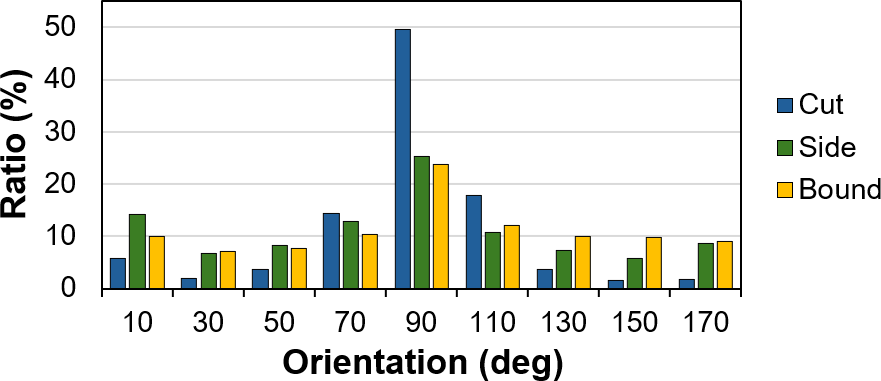}
    \caption{Distribution of instance orientations by class in 20-degree intervals around the given values. The angles are measured between the longest oriented-bounding-box side and the horizontal axis.}
    \label{fig:orientation}
\end{figure}

As an addition to the dataset analysis, this section provides extended statistics regarding instance characteristics and distributions across the TimberVision dataset. Firstly, \cref{tab:log_components} shows the numbers of trunk components included in each subset. \cref{fig:instance_sizes} and \cref{fig:orientation} show the distributions of their sizes and orientations, respectively. As expected, \textit{Bound} instances are on average the smallest and \textit{Side} instances the largest class, with some of the latter even exceeding image dimensions if the trunk is located diagonally across the image. \cref{fig:instance_sizes} shows the elongated characteristic of \textit{Sides} and \textit{Bounds}, while \textit{Cuts} are closer to square shapes. On the other hand, there is a strong peak regarding orientations of \textit{Cut} instances. Since they are often viewed slightly from the side, they tend to form an upright oval projection in the image plane. \textit{Side} orientations are more uniformly distributed with a slight bias towards completely horizontal or vertical orientations, the latter resulting mainly from upright live trees in contrast to cut trees which are arbitrarily oriented in image space. \textit{Bounds} show a corresponding behaviour, as they form the end points of \textit{Side} instances.

\begin{figure}
    \centering
    \includegraphics[width=1\columnwidth]{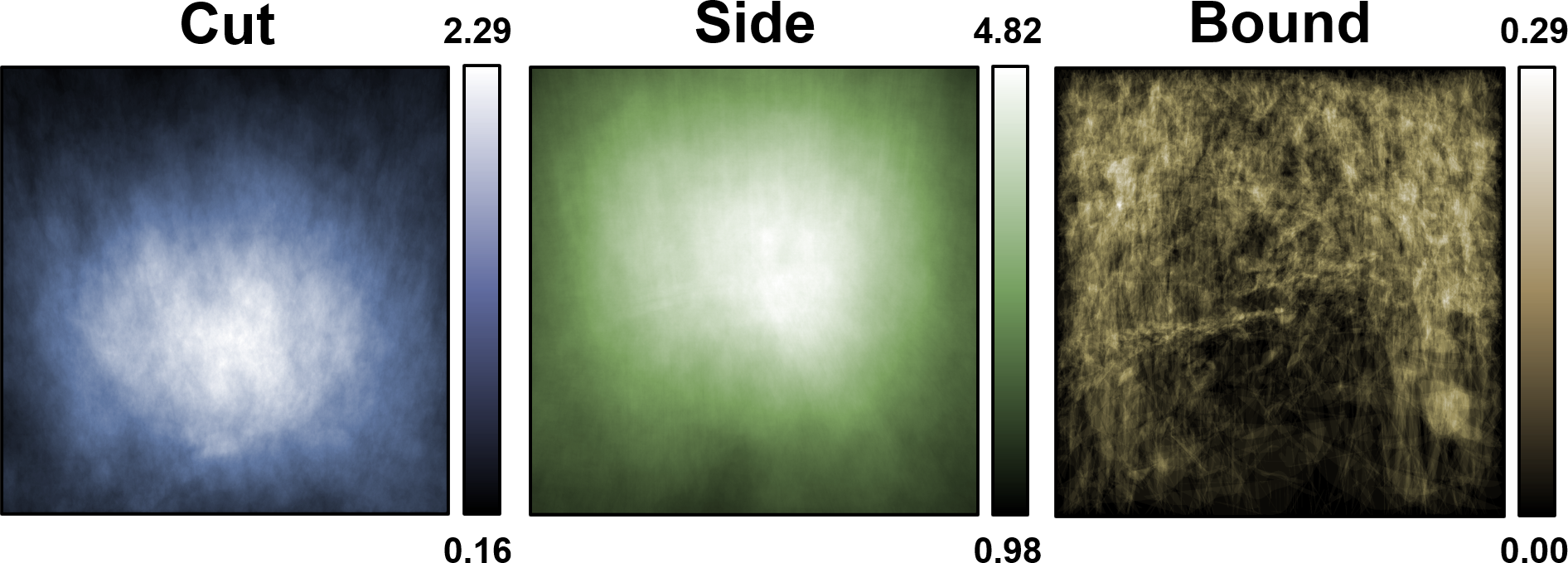}
    \caption{Heat maps illustrating the distribution of instance-segmentation masks within normalized image space for each class. Values are normalized between their respective minima and maxima, which are stated relative to the total numbers of instances for each class (i.e. a white area for the \textit{Side} class indicates that 4.82\% of \textit{Side} instances include this position).}
    \label{fig:heatmap}
\end{figure}

Furthermore, the heat maps in \cref{fig:heatmap} illustrate instance-mask distributions across normalized image space. \textit{Cut} and \textit{Side} instances appear in all positions across the area. However, since many images are captured with hand-held devices from an eye-level perspective, the former tend to be mainly in the lower central region, while the latter are often found in slightly higher positions. \textit{Bound} instances, on the other hand, are clustered along the image edges, as they are usually part of entire visible trunks. The distributions of all classes are largely symmetrical and closely related to realistic application scenarios. 

Overall, the statistics show that our data covers a wide range of relevant scene configurations. The characteristics of all classes closely match those to be expected in most target applications.

\begin{table}
    \begin{center}
        \begin{tabular}{ll|cccc}
            \multicolumn{1}{l}{} &  & \textbf{Core} & \textbf{Load} & \textbf{Harvest} & \textbf{Open} \\
            \hline
            \multirow{2}{*}{\textbf{OOD}} & \textbf{Cut} & 83.0 & 79.5 & 80.4 & 80.8 \\
            & \textbf{Side} & 59.8 & 66.3 & 47.1 & 45.0 \\
            \hline
            \multirow{2}{*}{\textbf{ISEG}} & \textbf{Cut} & 77.4 & 67.0 & 67.7 & 72.1 \\
            & \textbf{Side} & 61.2 & 69.7 & 45.3 & 45.4
        \end{tabular}
    \end{center}
    \caption{Model performance as \textit{mAP\textsuperscript{50-95}} for the classes \textit{Cut} and \textit{Side} for test and validation images of the \textbf{Core}, \textbf{Load}ing, \textbf{Harvest}ing and \textbf{Open}Source subsets. Results are reported for \textit{Large} models trained and evaluated on the same image resolution of 1024 pixels.}
    \label{tab:model_subset_results}
\end{table}

\begin{table*}
    \begin{center}
        \begin{tabular}{l|ccc|ccc|ccc}
            & \textbf{FP\textsubscript{ID}} $\downarrow$ & \textbf{FN\textsubscript{ID}} $\downarrow$ & \textbf{TP\textsubscript{ID}} $\uparrow$ & \textbf{MT} $\uparrow$ & \textbf{PT} $\uparrow$ & \textbf{ML} $\downarrow$ & \textbf{Misses} $\downarrow$ & \textbf{Switches} $\downarrow$ & \textbf{Frag.} $\downarrow$ \\
            \hline
            \textbf{ByteTrack} & 202 & 718 & 1,124 & 87 & 42 & 42 & 559 & 68 & \textbf{51} \\
            \textbf{Bot-SORT} & \textbf{181} & 696 & 1,146 & 87 & 42 & 42 & 560 & 66 & \textbf{51} \\
            \textbf{Optimized} & 194 & \textbf{681} & \textbf{1,161} & \textbf{89} & 43 & \textbf{39} & \textbf{534} & \textbf{63} & \textbf{51} \\
            \hline
            \textbf{Opt} $|$ 10fps & 245 & 806 & 1,036 & 82 & \textbf{48} & 41 & 606 & 129 & 65
        \end{tabular}
    \end{center}
    \caption{Additional Clear-Mot \cite{BeSt08} and ID \cite{RiSo16} metrics for all \textit{Tracking} sequences including 266 keyframes with 1,842 ground-truth annotations. \textbf{FP\textsubscript{ID}}, \textbf{FN\textsubscript{ID}} and \textbf{TP\textsubscript{ID}} denote false positives, false negatives and true positives according to the ID metric. \textbf{MT}, \textbf{PT} and \textbf{ML} denote the numbers of mostly tracked, partly tracked and mostly lost objects, respectively.}
    \label{tab:mot_extended_results}
\end{table*}

\section{Extended quantitative evaluation}
The following tables provide further details regarding the quantitative evaluation of ablation and detection experiments. \cref{tab:model_subset_results} shows individual detection results of the subsets comprising \textit{Base} in Tab. 3 of the paper to differentiate model performance for specific types of input data and scenarios. As opposed to the overall evaluation, the validation images are included in addition to the test set to achieve a representative number of samples for each subset. \textit{Core} constitutes the largest part of the test set and is therefore most closely related to the overall results. The \textit{Cut} class is consistently well detected across all subsets, while \textit{Side} achieves the best results for \textit{Loading} scenarios, but does not perform as well for \textit{Harvesting} and \textit{OpenSource} data. This is consistent with its performance on the \textit{TimberSeg*} subset presented in the paper, which commonly features these kinds of scenes. A possible reason for this behaviour, apart from the lower number of samples for harvesting scenarios, might be that the corresponding scenes often contain a high number of live trees in the far background, which are inconsistently detected and especially challenging for detection approaches in general.

Furthermore, detailed experimental results of our class ablations for oriented object detection and instance segmentation, which serve as the basis for Fig. 6 in the main paper, are listed in \cref{tab:class_ablation}. Additionally, the ablation of both learning tasks is shown in \cref{tab:full_ablation} and forms the basis for Fig. 5 in the main paper.

\section{Extended fusion and tracking results}
\begin{table}
    \begin{center}
        \begin{tabular}{l|ccc}
             & \textbf{Precision} & \textbf{Recall} & \textbf{mAP\textsuperscript{50-95}} \\
            \hline
            \textbf{Base} & 84.3 & 72.9 & 57.5 \\
            \textbf{TimberSeg} & - & 50.8 & - \\
        \end{tabular}
    \end{center}
    \caption{OBB accuracy for fused trunks on our test set (\textit{Base}) and the original TimberSeg dataset. Since the latter does not include annotations for all trunk instances, only recall is applicable.}
    \label{tab:fusion_eval}
\end{table} 

In addition to the fusion results presented in the main paper, \cref{tab:fusion_eval} gives an idea of the models' generalization capability using the original TimberSeg dataset. Since fusion results are entire \textit{Trunk} instances, we can use the provided annotations for this class, but still only compare recall, as they do not cover all visible instances (see \cref{fig:anno_comparison}). The performance drop is consistent with the one for our selected and newly annotated subset presented in Tab. 3 of the main paper. However, according to \cite{FoGa22}, even models trained and tested only on splits of the 220 TimberSeg images do not yield recalls beyond 65.2\% or \textit{mAP} scores beyond 57.5 for the same class, proving the challenging nature of the data.

\begin{figure}
    \centering
    \includegraphics[width=1\columnwidth]{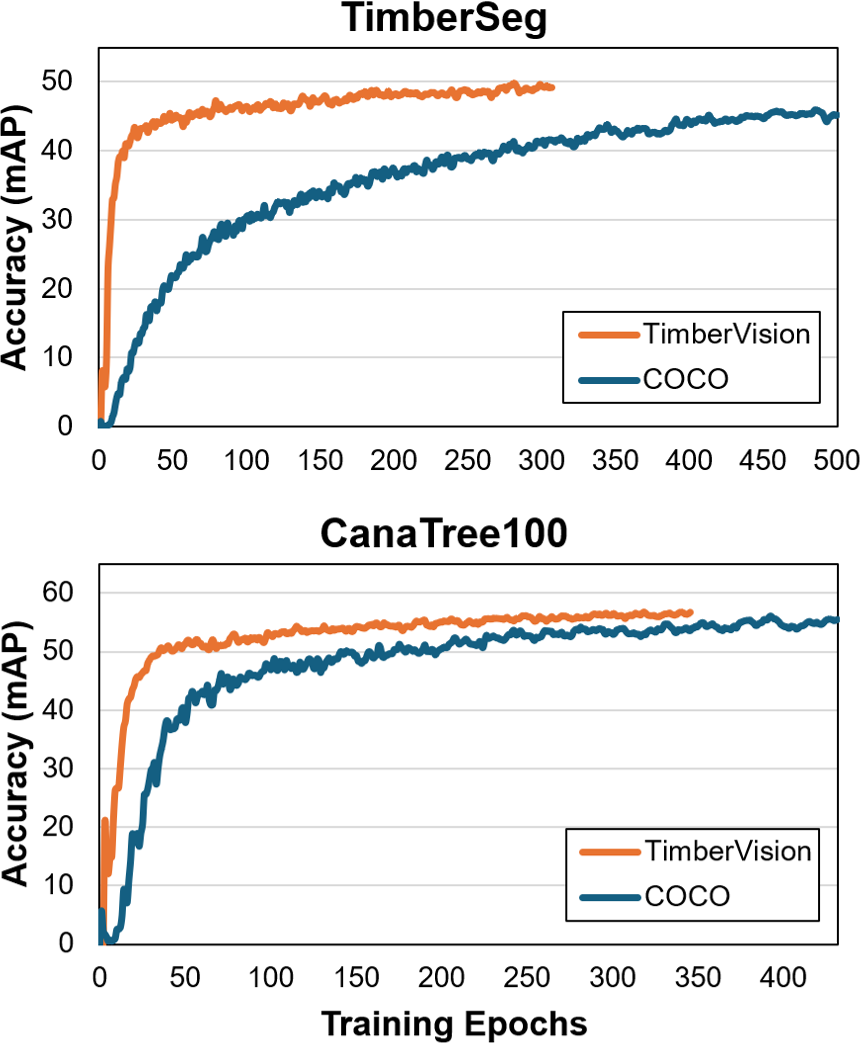}
    \caption{Validation accuracy after each training epoch when fine-tuning on TimberSeg and CanaTree100 with models pre-trained on MS COCO and our TimberVision dataset. Experiments are based on \textit{Large} model architectures with an input size of 1024. \textit{mAP} scores are derived from five-fold validation experiments.}
    \label{fig:finetuning}
\end{figure}

Fusion results serve as an input for multi-object tracking, for which we list additional MOT metrics in \cref{tab:mot_extended_results}. Fine-tuning experiments on TimberSeg and CanaTree100 with models pre-trained on our TimberVision dataset and MS COCO are illustrated by \cref{fig:finetuning}, which shows \textit{mAP} scores on the validation set for each training epoch of the experiments described in Tab. 5 of the main paper. As discussed, training times can be significantly reduced when using our models as basis for fine-tuning datasets of similar domains.

\begin{table*}
    \begin{center}
        \begin{tabular}{c|c|ccc|cc|c|cccc|cc}
            \multicolumn{1}{c}{} & \multicolumn{1}{c}{} & \multicolumn{6}{|c|}{\textbf{Oriented Object Detection}} & \multicolumn{6}{c}{\textbf{Instance Segmentation}} \\
            \multicolumn{1}{c}{} & \multicolumn{1}{|c}{\textbf{Size}} & \multicolumn{1}{|c}{\textbf{C}} & \textbf{S} & \textbf{B} & \textbf{C} & \textbf{S} & \textbf{T} & \textbf{C\textsubscript{Box}} & \textbf{S\textsubscript{Box}} & \textbf{C\textsubscript{Mask}} & \textbf{S\textsubscript{Mask}} & \textbf{T\textsubscript{Box}} & \textbf{T\textsubscript{Mask}} \\            
            \hline
            \multirow{2}{*}{\textbf{n}} & \textbf{768} & 76.5 & 22.6 & 49.7 & 76.7 & 49.1 & 55.9 & 75.5 & 55.5 & 69.8 & 49.0 & 64.6 & 59.1 \\
            & \textbf{1024} & 77.8 & 22.3 & 49.3 & 77.7 & 50.1 & 56.8 & 77.5 & 56.7 & 72.8 & 50.3 & 64.7 & 59.6 \\
            \hline
            \multirow{2}{*}{\textbf{m}} & \textbf{768} & 79.8 & 25.7 & 54.7 & 79.8 & 54.7 & 61.2 & 78.7 & 62.0 & 72.9 & 56.5 & 69.2 & 65.4 \\
            & \textbf{1024} & 80.8 & 27.0 & 55.9 & 81.1 & 56.0 & 62.7 & 80.0 & 62.0 & 75.5 & 56.6 & 70.5 & 66.5 \\
            \hline
            \multirow{2}{*}{\textbf{x}} & \textbf{768} & 80.8 & 28.7 & 56.6 & 81.0 & 56.6 & 63.8 & 79.5 & 63.1 & 73.9 & 58.7 & 70.6 & 67.0 \\
            & \textbf{1024} & \textbf{81.9} & \textbf{30.5} & \textbf{58.4} & \textbf{82.1} & \textbf{58.4} & \textbf{65.3} & \textbf{80.7} & \textbf{64.1} & \textbf{75.8} & \textbf{59.3} & \textbf{71.5} & \textbf{68.3} \\
        \end{tabular}
        \caption{Complete results of class ablation experiments for the model capacities \textit{Nano} (\textbf{n}), \textit{Medium} (\textbf{m}) and \textit{X-Large} (\textbf{x}) and different image sizes for the classes \textit{\textbf{C}ut}, \textit{\textbf{S}ide}, \textit{\textbf{B}ound} and \textit{\textbf{T}runk}. All scores are given as \textit{mAP\textsuperscript{50-95}} on the test set. In the case of instance segmentation, scores are reported separately for the \textit{Box} and \textit{Mask} stages.}
    \label{tab:class_ablation}
    \end{center}
\end{table*}

\begin{table*}
    \begin{center}
        \begin{tabular}{c|c|ccc|c|ccc|ccc|c}
            \multicolumn{1}{l}{} & \multicolumn{1}{l}{} & \multicolumn{4}{|c}{\textbf{Oriented Object Detection}} & \multicolumn{7}{|c}{\textbf{Instance Segmentation}} \\      
            \multicolumn{1}{c}{} & \multicolumn{1}{|c}{\textbf{Size}} & \multicolumn{1}{|c}{\textbf{C}} & \textbf{S} & \textbf{\bm{$\varnothing$}} & \textbf{t} & \textbf{C\textsubscript{Box}} & \textbf{S\textsubscript{Box}} & \textbf{\bm{$\varnothing$}\textsubscript{Box}} & \textbf{C\textsubscript{Mask}} & \textbf{S\textsubscript{Mask}} & \textbf{\bm{$\varnothing$}\textsubscript{Mask}} & \textbf{t} \\
            \hline
            \multirow{5}{*}{\textbf{n}} & \textbf{640} & 74.7 & 47.5 & 61.1 & 1.8 & 73.7 & 53.8 & 63.8 & 67.2 & 47.2 & 57.2 & 1.9 \\
            & \textbf{768} & 76.7 & 49.1 & 62.9 & 2.0 & 75.5 & 55.5 & 65.5 & 69.8 & 49.0 & 59.4 & 2.3 \\
            & \textbf{896} & 77.5 & 49.6 & 63.6 & 2.2 & 76.3 & 56.1 & 66.2 & 71.2 & 50.1 & 60.7 & 2.6 \\
            & \textbf{1024} & 77.7 & 50.1 & 63.9 & 2.6 & 77.5 & 56.7 & 67.1 & 72.8 & 50.3 & 61.6 & 3.2 \\
            & \textbf{1152} & 78.5 & 51.0 & 64.8 & 3.1 & 78.4 & 56.5 & 67.5 & 73.6 & 50.6 & 62.1 & 3.9 \\
            \hline
            \multirow{5}{*}{\textbf{s}} & \textbf{640} & 77.2 & 50.6 & 63.9 & 2.5 & 76.3 & 57.1 & 66.7 & 69.6 & 50.7 & 60.2 & 3.0 \\
            & \textbf{768} & 77.9 & 51.9 & 64.9 & 3.4 & 77.5 & 58.2 & 67.9 & 71.4 & 52.3 & 61.9 & 4.2 \\
            & \textbf{896} & 79.1 & 52.7 & 65.9 & 4.3 & 78.5 & 58.6 & 68.6 & 73.1 & 52.8 & 63.0 & 5.3 \\
            & \textbf{1024} & 79.5 & 53.0 & 66.3 & 5.3 & 79.1 & 58.7 & 68.9 & 74.1 & 53.0 & 63.6 & 6.7 \\
            & \textbf{1152} & 79.9 & 53.4 & 66.7 & 6.5 & 79.3 & 59.2 & 69.3 & 75.1 & 53.6 & 64.4 & 8.2 \\
            \hline
            \multirow{5}{*}{\textbf{m}} & \textbf{640} & 78.4 & 53.8 & 66.1 & 4.8 & 77.7 & 60.1 & 68.9 & 71.3 & 54.6 & 63.0 & 5.7 \\
            & \textbf{768} & 79.8 & 54.7 & 67.3 & 6.9 & 78.7 & 62.0 & 70.4 & 72.9 & 56.5 & 64.7 & 8.3 \\
            & \textbf{896} & 80.2 & 55.7 & 68.0 & 9.2 & 79.7 & 61.8 & 70.8 & 74.4 & 56.9 & 65.7 & 11.1 \\
            & \textbf{1024} & 81.1 & 56.0 & 68.6 & 11.8 & 80.0 & 62.0 & 71.0 & 75.5 & 56.6 & 66.1 & 14.3 \\
            & \textbf{1152} & 81.4 & 55.9 & 68.7 & 14.4 & 80.0 & 62.4 & 71.2 & 75.8 & 56.9 & 66.4 & 17.4 \\
            \hline
            \multirow{5}{*}{\textbf{l}} & \textbf{640} & 79.7 & 55.4 & 67.6 & 7.9 & 78.4 & 62.0 & 70.2 & 71.6 & 56.8 & 64.2 & 9.4 \\
            & \textbf{768} & 80.5 & 56.1 & 68.3 & 11.4 & 79.4 & 63.6 & 71.5 & 73.6 & 58.3 & 66.0 & 13.6 \\
            & \textbf{896} & 81.3 & 57.4 & 69.4 & 15.4 & 80.2 & 63.4 & 71.8 & 75.1 & 58.6 & 66.9 & 18.3 \\
            & \textbf{1024} & 81.8 & 58.0 & 69.9 & 19.7 & 80.4 & \textbf{64.2} & 72.3 & 76.0 & 58.9 & 67.5 & 23.3 \\
            & \textbf{1152} & 82.2 & 57.9 & 70.1 & 24.2 & 80.2 & 64.1 & 72.2 & 75.6 & 58.5 & 67.1 & 28.8 \\
            \hline
            \multirow{5}{*}{\textbf{x}} & \textbf{640} & 80.0 & 55.5 & 67.8 & 12.2 & 78.4 & 62.3 & 70.4 & 71.6 & 56.8 & 64.2 & 14.5 \\
            & \textbf{768} & 81.0 & 56.6 & 68.8 & 17.8 & 79.5 & 63.1 & 71.3 & 73.9 & 58.7 & 66.3 & 21.2 \\
            & \textbf{896} & 82.0 & 58.2 & 70.1 & 24.1 & 80.1 & 63.9 & 72.0 & 74.7 & \textbf{59.4} & 67.1 & 28.2 \\
            & \textbf{1024} & 82.1 & 58.4 & 70.3 & 30.6 & 80.7 & 64.1 & 72.4 & 75.8 & 59.3 & 67.6 & 36.1 \\
            & \textbf{1152} & \textbf{82.3} & \textbf{59.2} & \textbf{70.8} & 37.6 & \textbf{80.9} & 64.1 & \textbf{72.5} & \textbf{76.4} & 59.2 & \textbf{67.8} & 44.1 \\
        \end{tabular}
    \caption{Complete results of oriented-object-detection and instance-segmentation experiments for the model capacities \textit{Nano} (\textbf{n}), \textit{Small} (\textbf{s}), \textit{Medium} (\textbf{m}), \textit{Large} (\textbf{l}) and \textit{X-Large} (\textbf{x}) and different input sizes. The \textit{mAP\textsuperscript{50-95}} scores on the test set are listed for the classes \textit{\textbf{C}ut} and \textit{\textbf{S}ide} as well as their average (\textbf{\bm{$\varnothing$}}) along with mean inference time (\textbf{t}) in milliseconds. In the case of instance segmentation, scores are listed for the \textit{Box} and \textit{Mask} stages separately.}
    \label{tab:full_ablation}
    \end{center}
\end{table*}

\section{Extended qualitative results}
\begin{figure}
    \centering
    \includegraphics[width=1\columnwidth]{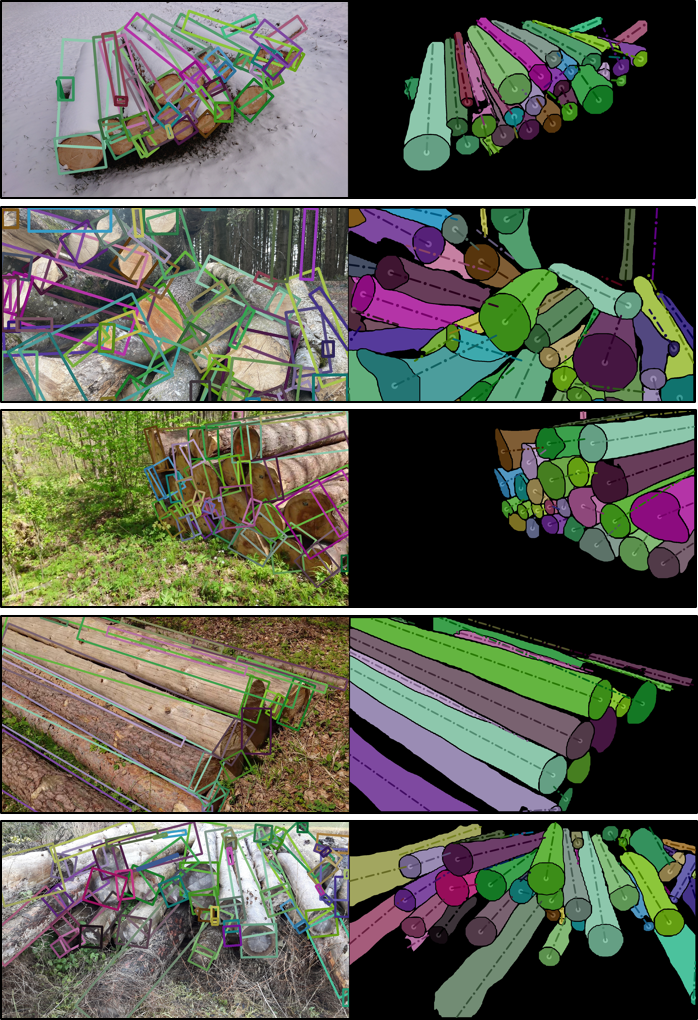}
    \caption{Additional qualitative results on the test split of the \textit{Core} subset recorded in forests and other outdoor locations.}
    \label{fig:results_core}
\end{figure}

\begin{figure}
    \centering
    \includegraphics[width=1\columnwidth]{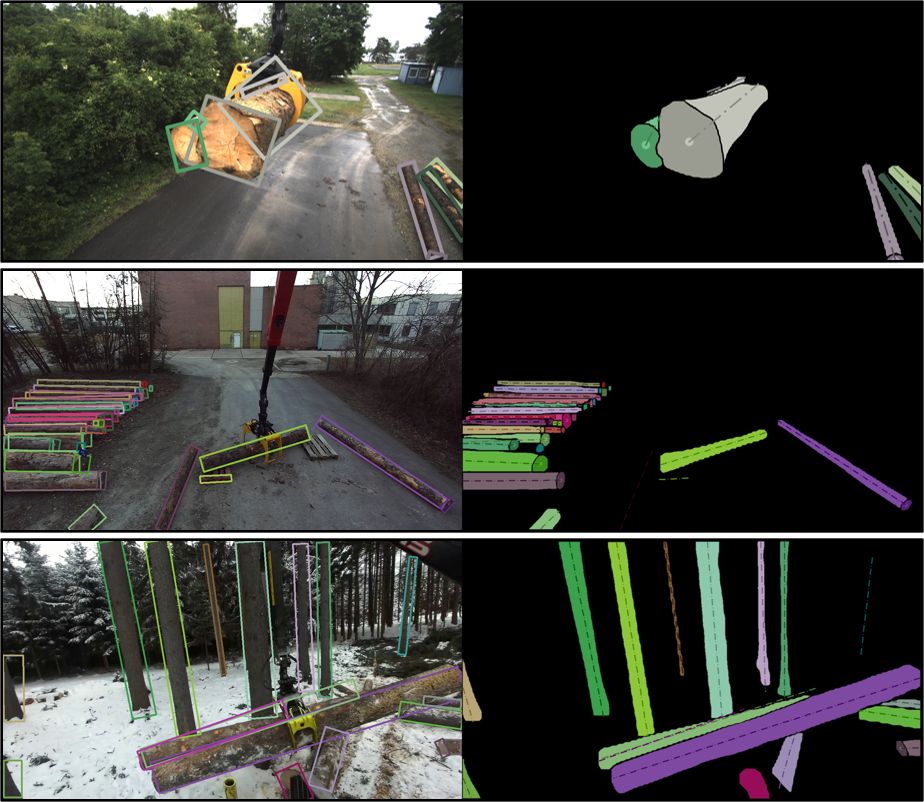}
    \caption{Additional qualitative results on the test splits of the \textit{Loading} and \textit{Harvesting} subsets depicting realistic application scenarios.}
    \label{fig:results_loading_harvesting}
\end{figure}

\begin{figure}
    \centering
    \includegraphics[width=1\columnwidth]{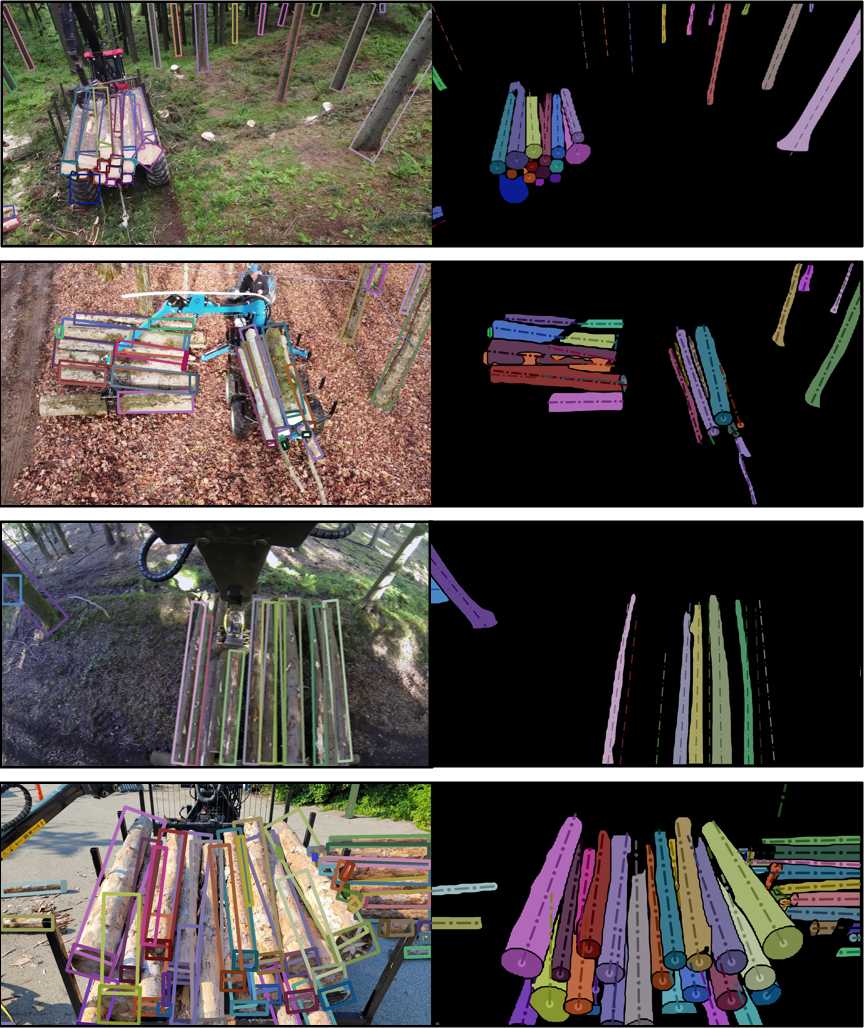}
    \caption{Additional qualitative results on the TimberSeg dataset \cite{FoGa22} demonstrating the generalization capability of our approach.}
    \label{fig:results_timberseg}
\end{figure}

\begin{figure}
    \centering
    \includegraphics[width=1\columnwidth]{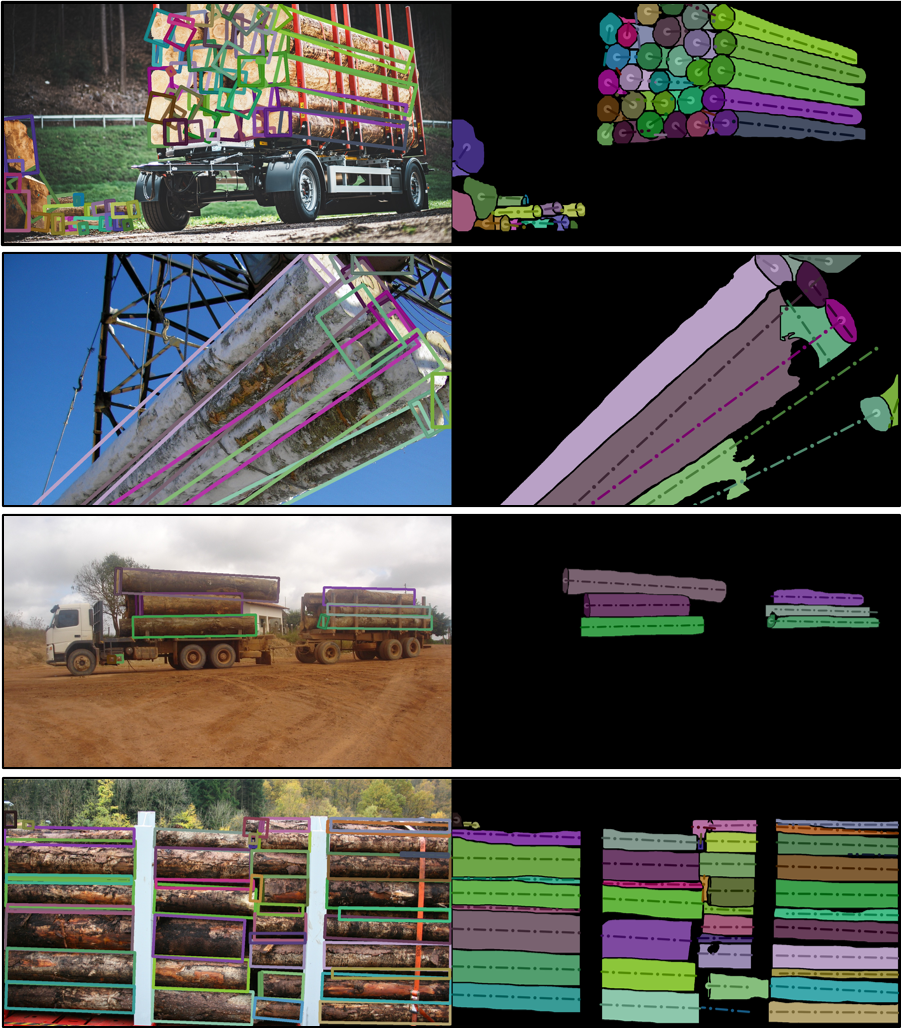}
    \caption{Additional qualitative results on the test split of the \textit{OpenSource} subset with complementary scenarios to the main data from public sources.}
    \label{fig:results_opensource}
\end{figure}

\begin{figure}
    \centering
    \includegraphics[width=1\columnwidth]{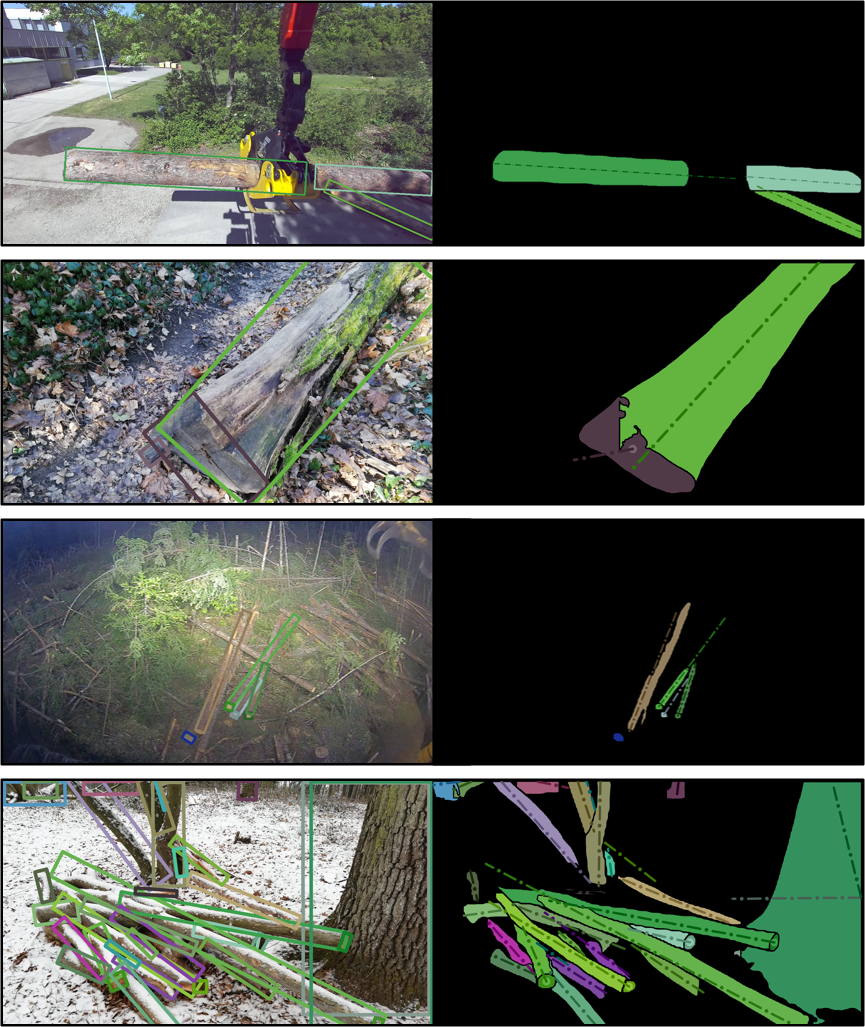}
    \caption{Additional qualitative results showing limitations on our test set and the TimberSeg dataset \cite{FoGa22}.}
    \label{fig:results_cornercases}
\end{figure}

To further illustrate our discussion of results and potential application scenarios, we show extended qualitative results, clustered by their subsets of the TimberVision dataset. None of the listed images were included during training or validation. In addition to representative results on the \textit{Core} subset (\cref{fig:results_core}) and in typical \textit{Loading} and \textit{Harvesting} scenarios (\cref{fig:results_loading_harvesting}), we demonstrate the generalization potential of our approach on samples of the \textit{TimberSeg} dataset \cite{FoGa22} (\cref{fig:results_timberseg}) and \textit{OpenSource} images (\cref{fig:results_opensource}). This is complemented by images from all subsets and TimberSeg showing selected corner cases and limitations (\cref{fig:results_cornercases}) to identify challenging scenarios and potentials for improvement. As discussed in the main paper, especially images in low-light conditions and trunks triggering multiple detections due to large occlusions need further investigation during future iterations of the dataset.

% %\clearpage

% {\small
% \bibliographystyle{ieee_fullname}
% \bibliography{references}
% }

% \end{document}

\end{document}